\documentclass[10pt,twocolumn,letterpaper]{article}

\usepackage{cvpr}              % To produce the CAMERA-READY version
\usepackage[dvipsnames, table]{xcolor}

\definecolor{cvprblue}{rgb}{0.21,0.49,0.74}
\usepackage[pagebackref,breaklinks,colorlinks,citecolor=cvprblue]{hyperref}

\usepackage{makecell, multirow, tabularx}
\usepackage{ragged2e} 
\usepackage{threeparttable}
\usepackage{paralist}
\usepackage{enumitem}

\usepackage{pifont}
\usepackage[dvipsnames]{xcolor}

\usepackage{silence}
\makeatletter
\robustify\@latex@warning@no@line
\makeatother
\usepackage{authblk}

\usepackage{stackengine}

\makeatletter
\newcommand\email[2][]%
   {\newaffiltrue\let\AB@blk@and\AB@pand
      \if\relax#1\relax\def\AB@note{\AB@thenote}\else\def\AB@note{\relax}%
        \setcounter{Maxaffil}{0}\fi
      \begingroup
        \let\protect\@unexpandable@protect
        \def\thanks{\protect\thanks}\def\footnote{\protect\footnote}%
        \@temptokena=\expandafter{\AB@authors}%
        {\def\\{\protect\\\protect\Affilfont}\xdef\AB@temp{#2}}%
         \xdef\AB@authors{\the\@temptokena\AB@las\AB@au@str
         \protect\\[\affilsep]\protect\Affilfont\AB@temp}%
         \gdef\AB@las{}\gdef\AB@au@str{}%
        {\def\\{, \ignorespaces}\xdef\AB@temp{#2}}%
        \@temptokena=\expandafter{\AB@affillist}%
        \xdef\AB@affillist{\the\@temptokena \AB@affilsep
          \AB@affilnote{}\protect\Affilfont\AB@temp}%
      \endgroup
       \let\AB@affilsep\AB@affilsepx
}
\makeatother

\def\paperID{9499} % *** Enter the Paper ID here
\def\confName{CVPR}
\def\confYear{2024}

\title{Multi-modal Instruction Tuned LLMs with Fine-grained Visual Perception}

\author[1,2]{Junwen He\thanks{Work done during internship at DAMO Academy.}}
\author[1]{Yifan Wang}
\author[1]{Lijun Wang\thanks{Corresponding author}}
\author[1]{Huchuan Lu}
\author[2]{Jun-Yan He}
\author[2]{\\ Jin-Peng Lan}
\author[2]{Bin Luo}
\author[2]{Xuansong Xie}

\affil[1]{Dalian University of Technology}
\affil[2]{DAMO Academy, Alibaba Group}
\email{\ttfamily \small {junwen.he@mail.dlut.edu.cn, \{wyfan, ljwang, lhchuan\}@dlut.edu.cn}}
\email{\ttfamily \small {\{leyuan.hjy, lanjinpeng.ljp, luwu.lb\}@alibaba-inc.com, xingtong.xxs@taobao.com }}

\begin{document}
\maketitle
\begin{abstract}

Multimodal Large Language Model (MLLMs) leverages Large Language Models as a cognitive framework for diverse visual-language tasks.
%Recent efforts have been made to make it spatial-aware, which enabling perceiving and grounding positional descriptions (\emph{e.g.}, bounding boxes) to the visual world.
Recent efforts have been made to equip MLLMs with visual perceiving and grounding capabilities.
However, there still remains a gap in providing fine-grained pixel-level perceptions and extending interactions beyond text-specific inputs.
% In this work, we propose {\bf{AnyRef}}, a general MLLM model capable of generating pixel-wise object perceptions as well as natural language descriptions from multi-modality references (\emph{e.g.}, texts, boxes, images or audios).
In this work, we propose {\bf{AnyRef}}, a general MLLM model that can generate pixel-wise object perceptions and natural language descriptions from multi-modality references, such as texts, boxes, images, or audio.
This innovation empowers users with greater flexibility to engage with the model beyond textual and regional prompts, without modality-specific designs.
Through our proposed refocusing mechanism, the generated grounding output is guided to better focus on the referenced object, implicitly incorporating additional pixel-level supervision.
% Our proposed refocusing mechanism guides the generated grounding output to prioritize the referenced object, implicitly incorporating additional pixel-level supervision.
This simple modification utilizes attention scores generated during the inference of LLM, eliminating the need for extra computations while exhibiting performance enhancements in both grounding masks and referring expressions.
With only publicly available training data, our model achieves state-of-the-art results across multiple benchmarks, including diverse modality referring segmentation and region-level referring expression generation.
Code and models are available at \url{https://github.com/jwh97nn/AnyRef}
\vspace{-0.3cm}

\end{abstract}    
\section{Introduction}
\label{sec:intro}

% introduce llm+mllm. other modality imagebind/languagebind.
% regional mllm. kosmos/shikra/gpt4roi/.
%Large language models (LLMs) have garnered widespread influence across various domains, and advancements have been achieved by augmenting LLMs with visual perception modules to bridge the gap between vision and language tasks \cite{instructblip, liu2023llava, li2023otter, zhu2023minigpt4}, thereby transforming them into Multimodal Large Language Models (MLLMs).
%Advancing a step further, spatial and regional information is integrated into MLLMs to enhance their capacity for perceiving and grounding object descriptions through user-defined formats (\emph{e.g.}, coordinates, bounding boxes, etc.) \cite{peng2023kosmos, chen2023shikra, zhang2023gpt4roi}, surpassing the confines of textual responses alone.

Large language models (LLMs) have garnered widespread influence across various domains, and advancements have been achieved by augmenting LLMs with visual perception modules to bridge the gap between vision and language tasks \cite{instructblip, liu2023llava, li2023otter, zhu2023minigpt4}, thereby transforming them into Multimodal Large Language Models (MLLMs).
Most recent research aims to further endow MLLMs with finer-grained visual understanding abilities, like visual grounding and referring expression generation, through user-defined formats (\emph{e.g.}, coordinates, bounding boxes, etc.) \cite{peng2023kosmos, chen2023shikra, zhang2023gpt4roi}, surpassing the confines of textual responses alone.

% \vspace{-0.5cm}
\begin{figure}[t]
    \centering
    \includegraphics[width=\columnwidth]{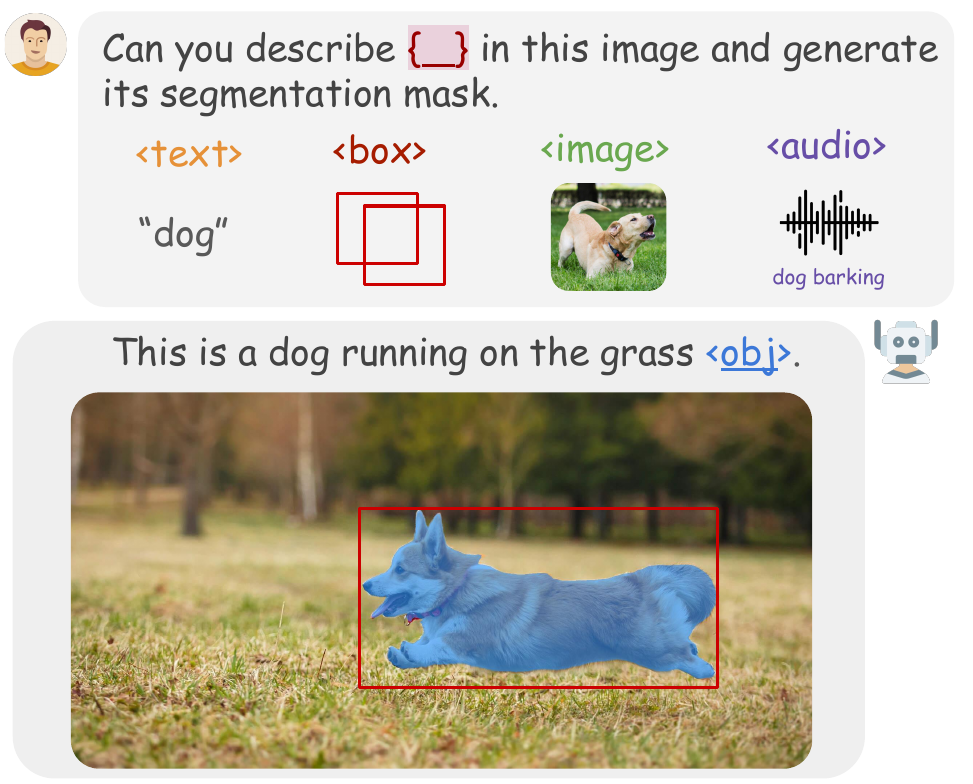}
    \caption{
    {\bf{Multi-modality Referring Segmentation}} and {\bf{Expression Generation}} with {\bf{AnyRef}}. Our model possesses the capacity to generate natural language descriptions as well as pixel-wise grounding masks for the referred object. It accommodates various referring modalities such as {\bf{\color{orange}text}}, {\bf{\color{BrickRed}bounding boxes}}, {\bf{\color{ForestGreen}images}} and {\bf{\color{Fuchsia}audio}}, enabling more flexible user interactions.
    }
    \label{intropdf}
    \vspace{-0.3cm}
\end{figure}
% \vspace{-0.5cm}

% Even though these models show promising grounding capability, they can only provide coarse-grained localization (bounding boxes), instead of pixel-level perceptions.
% The most recent work \cite{lai2023lisa} explores equipping the MLLM with segmentation models, to generate binary segmentation masks from textual descriptions or complex reasoning text queries.
% However, it is limited to textual referring instructions only, which restricts the potential applicability of MLLMs in multi-modality interactive tasks, such as region or audio understandings.
% SAM, SEEM
% Considering the above limitations, we introduce XXX, a general MLLM with capabilities of pixel-level object grounding and region-aware expression generation from multi-modality references, including texts, bounding boxes, images and audios.
Despite the encouraging results demonstrated by existing MLLMs in grounding linguistic expressions to visual scenes, their capacity for precise localization remains restricted to coarse-grained levels (bounding boxes), falling short of pixel-level perceptions (As illustrated in \cref{modeltable}).
The most recent work, as exemplified by \cite{lai2023lisa}, has focused on enhancing MLLMs by integrating segmentation models that generate binary segmentation masks based on textual descriptions.
However, this approach is constrained by its reliance solely on textual referring instructions, thereby limiting the versatility of MLLMs in various multimodal interaction scenarios, such as region-based referring or audio comprehension tasks.
The interactive segmentation model SEEM \cite{zou2023seem} attempts to receive audio inputs, but it turns audio into textural prompts with the off-the-shelf speech recognition model Whisper \cite{radford2023whisper}, so essentially it is still the textual references.

\newcommand{\cmark}{\textcolor{green!40!black}{\ding{51}}}
\newcommand{\xmark}{\textcolor{red}{\ding{55}}}

\begin{table*}[]
\centering
\resizebox{0.85\textwidth}{!}
{
\begin{tabular}{lcccccc}
% \hline
\toprule
\multirow{2}{*}{Method} & \multirow{2}{*}{Image} & \multicolumn{3}{c}{Referring Format} & \multirow{2}{*}{\begin{tabular}[c]{@{}c@{}}Pixel-level\\ Grounding\end{tabular}} & \multirow{2}{*}{\begin{tabular}[c]{@{}c@{}}End-to-End\\ Model\end{tabular}} \\
% \multirow{2}{*}{\begin{tabular}[c]{@{}c@{}}Multi-round\\ Conversation\end{tabular}} & \multirow{2}{*}{\begin{tabular}[c]{@{}c@{}}End-to-End\\ Model\end{tabular}} \\
% \cline{3-5}
\cmidrule{3-5}
 & & Region & Image* & Audio & & \\
% \hline
\midrule
LLaVA (NeurIPS-23) \cite{liu2023llava} & \cmark & \xmark & \xmark & \xmark & \xmark & \cmark \\
\rowcolor{gray!10}
BuboGPT (arXiv-23) \cite{zhao2023bubogpt} & \cmark & \xmark & \xmark & \cmark & \xmark  & \xmark \\
Vision-LLM (arXiv-23) \cite{wang2023visionllm} & \cmark & \cmark & \xmark & \xmark & \xmark & \cmark \\
DetGPT (arXiv-23) \cite{wang2023visionllm} & \cmark & \cmark & \xmark & \xmark & \xmark & \cmark \\
\rowcolor{gray!10}
KOSMOS-2 (arXiv-23) \cite{peng2023kosmos} & \cmark & \cmark & \xmark & \xmark & \xmark & \cmark \\
Shikra (arXiv-23) \cite{chen2023shikra} & \cmark & \cmark & \xmark & \xmark & \xmark & \cmark \\
\rowcolor{gray!10}
GPT4RoI (arXiv-23) \cite{zhang2023gpt4roi} & \cmark & \cmark & \xmark & \xmark & \xmark & \cmark \\
NExT-GPT (arXiv-23) \cite{wang2023seggpt} & \cmark & \xmark & \xmark & \cmark & \xmark & \cmark \\
\rowcolor{gray!10}
ASM (arXiv-23) \cite{wang2023asm} & \cmark & \cmark & \xmark & \xmark & \xmark & \cmark \\
LISA (arXiv-23) \cite{lai2023lisa} & \cmark & \xmark & \xmark & \xmark & \cmark  & \cmark \\
\rowcolor{orange!10}
{\bf{AnyRef}} (Ours) & \cmark & \cmark & \cmark & \cmark & \cmark & \cmark \\
% \hline
\bottomrule
\end{tabular}
}
\caption{{\bf{Comparisons of recent Multi-modal Large Language Models.}} The term \emph{Referring Format} emphasizes the acceptable modalities used for referencing, whereas \emph{Image*} indicates visual references derived from another image.}
\label{modeltable}
\vspace{-0.3cm}
\end{table*}

In light of the above observation, we propose {\bf{AnyRef}}, a novel multi-modal instruction-tuned LLM with fine-grained visual perception. As shown in \cref{modeltable}, {\bf{AnyRef}} advances existing MLLMs with the strong capability to perform pixel-level object grounding and generate region-aware expressions derived from references of diverse modalities, including text, bounding boxes, images, and audio inputs,  (See \cref{intropdf} as an example).
%
%To empower {\bf{AnyRef}} with the aforementioned capabilities, 
%To this end, we first unify the referring representations across various modalities and align them within the token space of LLMs.
%Taking into account all previously mentioned modalities, we extract their features to form the \emph{Unified Referring Representation}, which can be uniformly processed by the LLM, leveraging its ability of understanding and reasoning in generating the grounded output.
%This allows for flexible referring beyond merely textual descriptions, without modality-specific designs or modifications to the existing model.
%
To this end, we first propose a unified representation for referring across different modalities and map them to the token space of LLMs. We extract features from all the modalities mentioned above to form the \emph{Unified Referring Representation}, which can be processed uniformly by the LLM, utilizing its ability of understanding and reasoning in generating the grounded output. This enables flexible referring beyond textual descriptions, without requiring modality-specific designs or changes to the existing model.
%Additionally, the {\textless{\texttt{obj}}\textgreater} token is introduced to represent the querying segmentation mask, whose embedding will subsequently be employed as the input to the segmentation model \cite{kirillov2023segany} for decoding the referring segmentation masks.
% To empower the current MLLM \cite{liu2023llava} with the aforementioned capabilities, we introduce several additional tokens to the pre-defined vocabulary.
% Specifically, the obj token is employed to represent the querying segmentation mask, whose embedding will subsequently be extracted as the input to the segmentation model \cite{kirillov2023segany} for decoding the referring segmentation masks.
% Additionally, we use img\_ref, /img\_ref and aud\_ref, /aud\_ref to represent the beginning and end of the multi-modality references (bounding boxes/images and audios, respectively), with visual and audio features inserted between them.
% In this manner, this approach unifies the input reference format across diverse modalities, leveraging the LLM as a middleware.
% This allows for the flexible referring beyond merely textual descriptions, without modality-specific designs or modifications.

% framework for multi-modality referring

To perform pixel-level grounding with LLMs, a possible solution~\cite{lai2023lisa} is to trigger the segmentation action by generating a special token {\textless{\texttt{obj}}\textgreater}, whose embedding will be subsequently employed as the input to the segmentation model. As opposed to using coordinates sequence of polygons \cite{wang2023visionllm, chen2022pix2seq2} to represent segmentation results, the introduction of the {\textless{\texttt{obj}}\textgreater} token effectively simplifies pixel-level visual grounding.
%Incorporating the segmentation mask into a single pre-defined token in the vocabulary simplifies the generation of fine-grained perceptions, as opposed to multiple coordinate tokens as polygons \cite{wang2023visionllm, chen2022pix2seq2}.
Nevertheless, the embedding of the {\textless{\texttt{obj}}\textgreater}  token is confined in a fixed feature space, due to the nature of next token prediction, leading to limited representational capacity and thus inaccurate segmentation results.
To address this constraint, we propose a simple yet effective \emph{refocusing mechanism}, which takes into account the correlation between the grounded expression and the {\textless{\texttt{obj}}\textgreater}  token.
This mechanism utilizes attention scores to weight such correlation, enhancing the mask embedding with additional grounded embeddings, and since the attention scores are intermediate outputs of the self-attention layers, the additional computation introduced by the refocusing mechanism is minimal. 
%Furthermore, the generated grounding output before the {\textless{\texttt{obj}}\textgreater}  token will be implicitly guided by additional pixel-level supervisions, thereby enhancing the model's regional understanding capability.
Furthermore, the refocusing mechanism also provides a short-cut connection between the generated grounded expression and the segmentation results, allowing pixel-level labels to implicitly supervise the learning process of language expression generation, thereby enhancing the model's regional understanding capability.

To summarize, our contributions are threefold:
\begin{compactitem}
    \item We introduce {\bf{AnyRef}}, the first general MLLM capable of producing pixel-level object perceptions as well as region-aware referring descriptions. It adeptly accommodates multi-modality references including texts, bounding boxes, images or audio in a general manner, fostering more flexible interactions for users.
    \item We propose a simple yet effective \emph{refocusing mechanism} to enhance the grounded mask predictions, leveraging the correlations of generated tokens without incurring additional computational overhead, and concurrently yields improvements in regional expression referring.
    \item Thorough experiments conducted on multiple datasets demonstrate the efficacy of the proposed method, resulting in state-of-the-art performance across a diverse range of multi-modality tasks.
\end{compactitem}

Our model is built upon LLaVA-7B~\cite{liu2023llava}, which can be efficiently fine-tuned with 8 NVIDIA 32G V100 GPUs, making our method easily reproducible at a reasonable computational cost.

\section{Related Works}
\label{sec:rw}

\subsection{Multi-modal Large Language Model}

Multi-modal Large Language Models (MLLMs), built upon large language models (LLMs) as their foundations, extend their capabilities beyond traditional textual understanding to incorporate various modalities such as images, videos, and audio.
% In line with the concept of instruction tuning, Flamingo \cite{alayrac2022flamingo} initially employs visual feature inputs as prompts, and achieves remarkable results across various visual-language tasks, including image captioning and VQA.
Building upon the concept of instruction tuning, Flamingo \cite{alayrac2022flamingo} utilizes visual feature inputs as prompts, resulting in impressive performance across diverse visual-language tasks such as image captioning and visual question answering (VQA).
Subsequent models, includin BLIP-2 \cite{li2023blip2}, LLaVA \cite{liu2023llava}, InstructBLIP \cite{instructblip}, Otter \cite{li2023otter} and LLaMa-Adapter \cite{zhang2023llamaadapter}, utilize additional generated visual instruction-following data for better visual-language alignment, and demonstrate impressive multi-modal chat abilities.
% Recent works extend MLLMs with region-aware capacity to tackle localization tasks.

% \noindent{\bf{Spatial MLLMs.}}
Recent studies expand the capabilities of MLLMs to address localization tasks with region-aware functionalities.
% KOSMOS-2 \cite{peng2023kosmos} and VisionLLM \cite{wang2023visionllm} introduce additional location tokens to the vocabulary to convert coordinates into textual representations and input them into LLMs for region understanding, while Shikra \cite{chen2023shikra} represents coordinates in natural language form.
KOSMOS-2 \cite{peng2023kosmos} and VisionLLM \cite{wang2023visionllm} introduce additional location tokens to the vocabulary, enabling the conversion of coordinates into textual representations. These representations are then inputted into LLMs to enhance region understanding. On the other hand, Shikra \cite{chen2023shikra} represents coordinates directly in natural language form.
In contrast, GPT4RoI \cite{zhang2023gpt4roi} streamlines the process by employing RoI-aligned visual features without incorporating explicit positional information. 

% However, these models have limitations in producing fine-grained perceptions such as pixel-level masks and are restricted to textural descriptions and regions within the image for referring expressions.
Nevertheless, these models lack the capacity to produce fine-grained perceptions (\emph{e.g.}, pixel-level masks), and restrict their referring expressions to textural descriptions and regions within the image.
Our model, leveraging the best of both worlds, not only generates pixel-level grounding masks, but also accommodates a broader range of referring formats (\emph{e.g.}, visual reference from other images or audio) in a unified manner.

% regional: kosmos/shikra/gpt4roi/

\subsection{Referring Segmentation}

\noindent{\bf{Referring Expression Segmentation}} translates explicit textual descriptions into corresponding pixel-level segmentations, requiring a comprehensive understanding of both visual content and linguistic expression.
Recent methods including SAM \cite{kirillov2023segany}, X-Decoder \cite{zou2023xdecoder} and SEEM \cite{zou2023seem} unify multiple segmentation tasks within a single model, supporting various human interaction methods.
% While LISA \cite{lai2023lisa} leverages the exceptional capacity of LLM to reason and comprehend textural instructions, and generate masks from the SAM \cite{kirillov2023segany} decoder.
While LISA \cite{lai2023lisa} utilizes the powerful reasoning and comprehension abilities of LLMs to process textural instructions and generate masks through the SAM \cite{kirillov2023segany} decoder.

\noindent{\bf{Visual Referring Segmentation}} can be related to one/few-shot segmentation, where an example of a certain object with its corresponding mask is provided to segment the same object in the query image \cite{min2021hsnet, hong2022vat, zhang2022fptrans, wang2023painter, wang2023seggpt}.
Recently, CLIPSeg \cite{lueddecke22clipseg} builds upon the CLIP model to treat the example image as a visual prompt, which can generalize to novel forms of prompts. 
% And SEEM \cite{zou2023seem} propose a visual sampler to pool all kinds of image features into visual prompts, and shows powerful generalization capability to images of other domains.
Painter \cite{wang2023painter} and SegGPT \cite{wang2023seggpt} utilize in-context learning to perform general vision tasks using input task prompts.
% This comparably new task 

\noindent{\bf{Audio-Visual Segmentation}} aims to generate pixel-level masks for object(s) emitting sound, initially introduced in \cite{zhou2022avs}.
AVSegFormer \cite{gao2023avsegformer} innovatively incorporates learnable audio queries, enabling selective attention to relevant visual features.
Additionally, AUSS \cite{ling2023heartoseg} proposes unmixing self-supervised losses to bridge the gap between audio signals and visual semantics.
% While they utilize audio-specific designs to 

While these models have achieved satisfactory results in their respective domains, there is currently a gap in addressing all referring tasks within a single model.
% The above methods more or less use modality-specific or task-specific designs, which cannot simply generalize well beyond their own tasks.
Most of the aforementioned methods rely on modality-specific or task-specific designs, which may not generalize well beyond their intended tasks.
Our approach leverages the robust comprehension ability of LLMs to concurrently tackle all these tasks while preserving the region-level reasoning capacity.
Additionally, the \emph{refocusing mechanism} aids in enhancing region-level referring expression through implicit pixel-level supervisions.
\section{Methods}

\begin{figure*}[t]
    \centering
    \includegraphics[width=\textwidth]{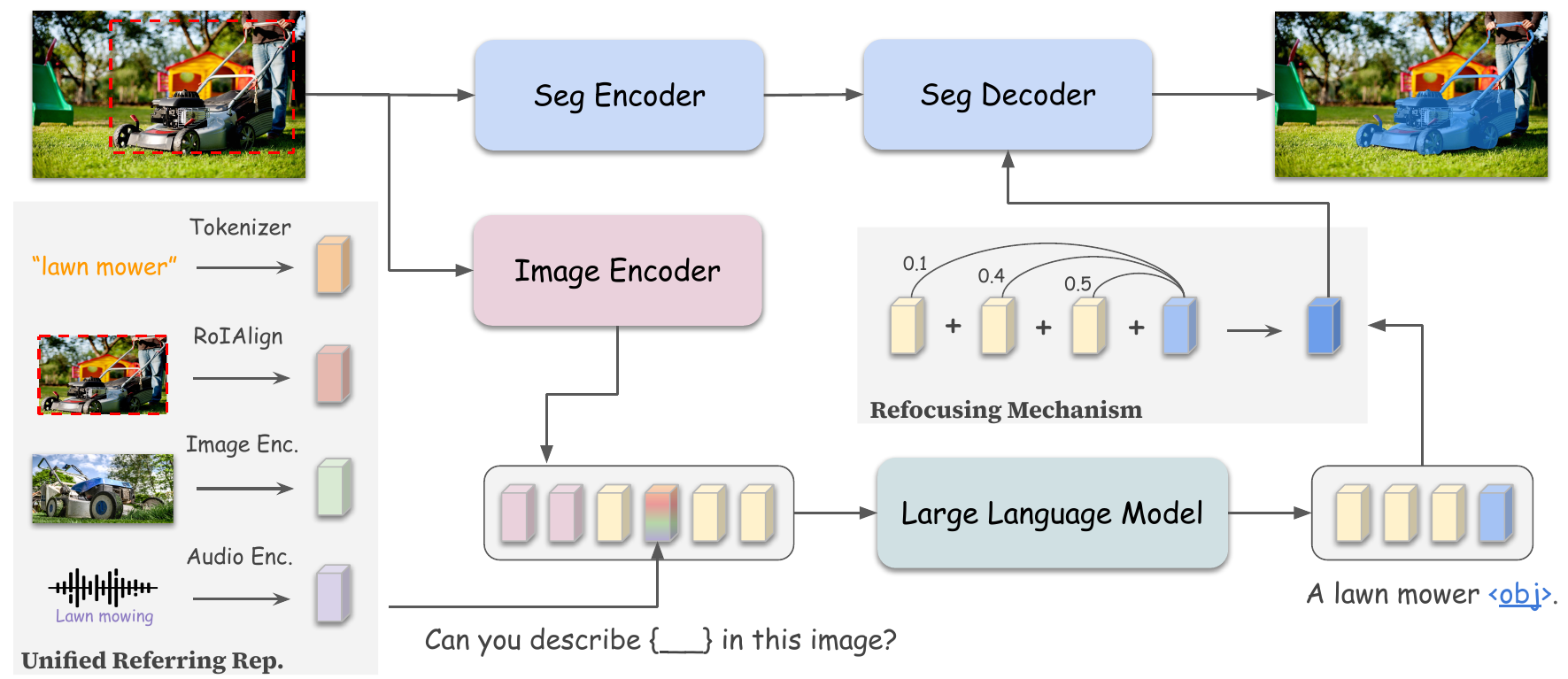}
    \caption{{\bf{Overall pipeline of AnyRef.}} Vision-language, audio-language projection and MLP layers are omitted for simplicity and clarity. The {\bf{Unified Referring Representation}} (\cref{unifiedreferringrepresentation}) receives references from diverse types of modalities and transforms them into embeddings aligned with the LLM. The {\bf{Refocusing Mechanism}} (\cref{refocusing}) enhances the embedding from the single {\textless{\texttt{obj}}\textgreater} token with grounded textural embeddings, thus providing a broader representational capacity.}
    \label{network}
    \vspace{-0.3cm}
\end{figure*}

The overall framework of {\bf{AnyRef}} comprises a vision encoder, multi-modal feature projection layers, a LLM, and a mask decoder, as illustrated in \cref{network}.
% These initial three components collectively constitute a multi-modality LLM, allowing for various reference formats and generating region-aware grounded textual responses.
These initial three components together form a multi-modality LLM, enabling support for various reference formats and generating region-aware grounded textual responses.
Additionally, a distinctive {\textless{\texttt{obj}}\textgreater} token is introduced to the vocabulary, which provides the input for the mask decoder through a refocusing mechanism, facilitating the generation of pixel-level perceptions.
% This token functions as the input for the mask decoder, facilitating the generation of pixel-level perceptions.
% serving as the input for the mask decoder to generate pixel-level perceptions.
% The overall pipeline is illustrated in \cref{network}.

\subsection{Model Architecture}

We adopt the pretrained ViT-L/14 from CLIP \cite{radford2021clip} as the vision encoder, and LLaMA-7B \cite{touvron2023llama} as our LLM.
For audio inputs, we choose the pretrained audio encoder from ImageBind \cite{girdhar2023imagebind} to extract audio features.
To connect multi-modality information beyond texts to the existing LLM, such as images and audio, we adopt vision-language and audio-language projection layers to project image and audio features to the language space.
% The input image and audio are converted into a fixed number of $16 \times 16$ and $3$ patch embeddings, respectively, and projected to the same dimension as word embeddings.
The input image is converted into a fixed number of $16 \times 16$ patch embeddings, while the audio is represented as $3$ patch embeddings.
Both the image and audio embeddings are then projected to the same dimension as word embeddings.
The LLM takes the interleaved embeddings in the same way as language tokens to generate outputs via an auto-regressive manner.
% The overall pipeline is illustrated in \cref{network}.

% \begin{figure*}[thp]
%     \centering
%     \includegraphics[width=\textwidth]{network.pdf}
%     \caption{{\bf{Overall pipeline of AnyRef.}} Vision-language, audio-language projection and MLP layers are omitted for simplicity and clarity. The {\bf{Unified Referring Representation}} (\cref{unifiedreferringrepresentation}) receives references from diverse types of modalities and transforms them into embeddings aligned with the LLM. The {\bf{Refocusing Mechanism}} (\cref{refocusing}) enhances the embedding from the single {\textless{\texttt{obj}}\textgreater} token with grounded textural embeddings, thus providing a broader representational capacity.}
%     \label{network}
% \end{figure*}

% \noindent{\textbf{Unified Referring Representation.}}
\subsubsection{Unified Referring Representation}
\label{unifiedreferringrepresentation}
To receive multi-modality referring prompts beyond texts, we convert them into fixed-sized tokens and \emph{quote} them between newly introduced special tokens.

For visual prompts including regional bounding boxes or visual examples from another image, we introduce {\textless{\texttt{img\_ref}}\textgreater} and {\textless{\texttt{/img\_ref}}\textgreater}, where visual features will be inserted in between.
Drawing inspiration from \cite{zhang2023gpt4roi}, we represent bounding boxes using extracted region-level features from RoIAlign \cite{he2017maskrcnn} with a fixed size of $4 \times 4$.
For processing image-level visual examples, we use the same CLIP vision encoder to extract visual features, which are then pooled to $4 \times 4$ as well.
To refer to them in the same way as textual descriptions, we build prompts such as: ``Can you provide a description of {\textless{\texttt{img\_ref}}\textgreater}{\textless{\texttt{img\_feat}}\textgreater}{\textless{\texttt{/img\_ref}}\textgreater} in this image?", where {\textless{\texttt{img\_feat}}\textgreater} will be replaced by the extracted visual features.

For audio prompts, we introduce {\textless{\texttt{aud\_ref}}\textgreater} and {\textless{\texttt{/aud\_ref}}\textgreater} for LLM to be aware of audio referring inputs, and the extracted audio features will be projected through audio-language projection layer and then inserted in between. 
And the audio prompted instruction will be built like: ``Can you segment the object that makes sound of {\textless{\texttt{aud\_ref}}\textgreater}{\textless{\texttt{aud\_feat}}\textgreater}{\textless{\texttt{/aud\_ref}}\textgreater} in this image?".
In this way, the referring representation from different modalities is unified, which can be treated the same way as language instructions and easily handled by the LLM.

\begin{figure*}[t]
    \centering
    \includegraphics[width=\textwidth]{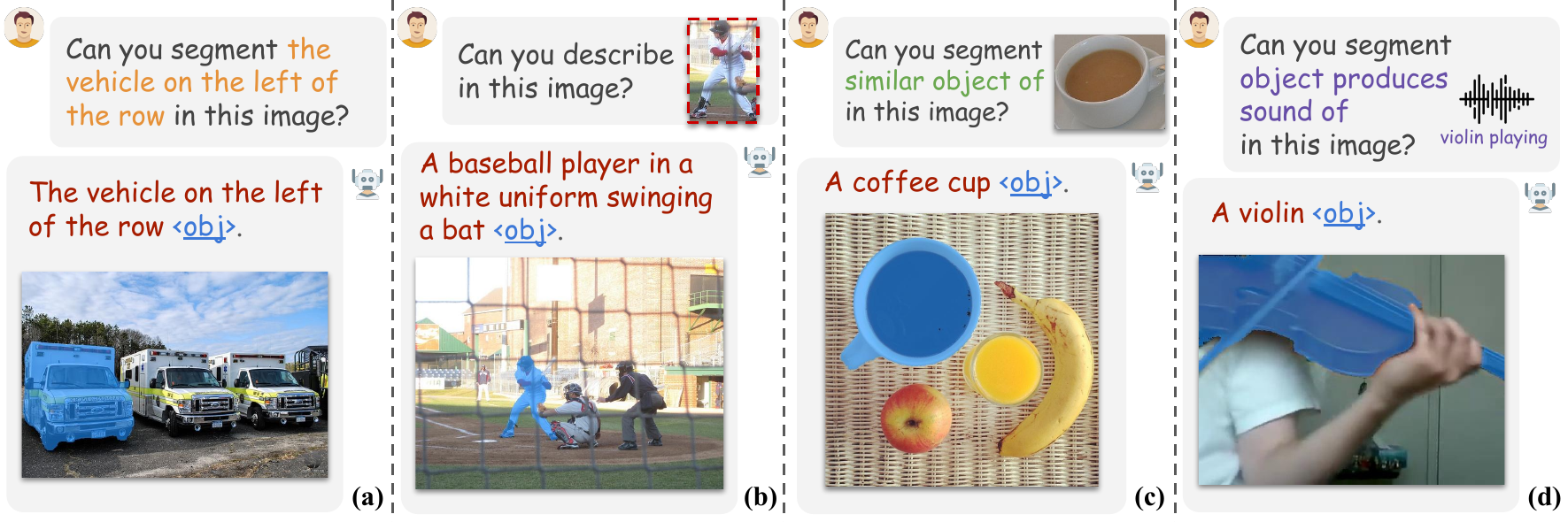}
    \caption{
    {\bf{Qualitative results of {\bf{AnyRef}}'s applicable capabilities}} on multiple tasks, including {\bf{(a)}} referring expression segmentation, {\bf{(b)}} region-level captioning and grounding, {\bf{(c)}} image-level referring segmentation and {\bf{(d)}} audio-visual segmentation. 
    {\bf{AnyRef}} demonstrates proficiency in generating both textual responses and pixel-level perceptions across diverse modality instructions.}
    \label{fig:task}
    \vspace{-0.3cm}
\end{figure*}

% \noindent{\textbf{Pixel-level Perception.}}
\subsubsection{Refocusing Mechanism}
\label{refocusing}
Inspired by \cite{lai2023lisa}, we employ another special token {\textless{\texttt{obj}}\textgreater} to succinctly represent the instance segmentation mask as an embedding.
This embedding $\boldsymbol{h}_{obj}$ is derived from the last-layer of LLM associated with the {\textless{\texttt{obj}}\textgreater} token.
It is then projected through an MLP layer $\gamma$, before being fed into the segmentation model $\mathcal{S}$.
Subsequently, the binary segmentation mask $M$ can be expressed mathematically as,
\begin{equation}
    M = \mathcal{S} \Big ( \gamma (\boldsymbol{h}_{obj}), \mathcal{V}_{seg} (\boldsymbol{x}_{img}) \Big ),
\end{equation}
where $\boldsymbol{x}_{img}$ indicates the input image, and $\mathcal{V}_{seg}$ denotes the vision encoder of the segmentation model.

However, since {\textless{\texttt{obj}}\textgreater} is a token in the LLM vocabulary, its representation will be limited in a fixed feature range, which will potentially limit its representational capacity and influence the decoded mask quality.
Therefore, we propose a \emph{refocusing mechanism} which augments the original mask embedding with grounded text embeddings.
The motivation behind is to explicitly force the final mask embedding to focus more on the referring or grounded object with its textural expression.
The updated mask embedding can be formulated as
\vspace{-0.3cm}
\begin{equation}
    \hat{\boldsymbol{h}}_{obj} = \boldsymbol{h}_{obj} + \lambda_{f} \sum_{i}^{i < obj} \bar{\boldsymbol{a}}_i \cdot \boldsymbol{h}_{i},
    \vspace{-0.3cm}
\end{equation}
% $i<obj \textless obj$
where $i \text{\textless} obj$ denotes the indices of output tokens before the {\textless{\texttt{obj}}\textgreater} token, $\bar{\boldsymbol{a}}_i$ indicates the normalized attention scores between the token $i$-th token and the {\textless{\texttt{obj}}\textgreater} token, and $\lambda_f=0.1$ controls the focusing weight of augmentation embeddings.
This approach enhances the mask embedding, providing a more adaptable feature range compared to the original, thereby expanding its representational capacity.

% \begin{figure*}[t]
%     \centering
%     \includegraphics[width=\textwidth]{task.pdf}
%     \caption{
%     {\bf{Qualitative results of {\bf{AnyRef}}'s applicable capabilities}} on multiple tasks, including {\bf{(a)}} referring expression segmentation, {\bf{(b)}} region-level captioning and grounding, {\bf{(c)}} image-level referring segmentation and {\bf{(d)}} audio-visual segmentation. 
%     {\bf{AnyRef}} demonstrates proficiency in generating both textual responses and pixel-level perceptions across diverse modality instructions.}
%     \label{fig:task}
% \end{figure*}

% \noindent{\textbf{Training Objectives.}}
\subsubsection{Training Objectives}
The model is trained in the end-to-end manner with a combination of text loss and mask loss. 
The text loss follows the next word prediction loss \cite{liu2023llava}, and the mask loss includes binary cross-entropy loss and dice loss \cite{milletari2016dice}, as
\begin{equation}
    \mathcal{L} = \lambda_{text} \mathcal{L}_{text} + \lambda_{bce} \mathcal{L}_{bce} + \lambda_{dice} \mathcal{L}_{dice},
\end{equation}
where we choose $\lambda_{text} = 1.0$, $\lambda_{bce}=2.0$ and $\lambda_{dice} = 0.5.$
Due to the \emph{refocusing mechanism}, tokens generated before the {\textless{\texttt{obj}}\textgreater} token can receive additional supervisory signals from pixel-level ground truth.
This mutual interaction can further benefit the vision-language understanding ability of \textbf{AnyRef}, given the interrelated nature of referring expressions and grounding masks.

\subsection{Implementation Details.}

% \noindent{\textbf{Training Setup.}}
\subsubsection{Training Setup}
Unless otherwise specified, we employ the pre-trained CLIP ViT-L/14 as the vision encoder, ImageBind-H \cite{girdhar2023imagebind} as the audio encoder, and LLaMa-7B as the LLM.
The vision-language projection layer is initialized from LLaVa \cite{liu2023llava}, while the audio-language projection layer is randomly initialized.
The word embeddings of newly introduced special tokens are initialized randomly.
Furthermore, the segmentation model utilizes the pre-trained SAM-H \cite{kirillov2023segany}.
The image resolution is $224 \times 224$ for MLLM and $1024 \times 1024$ by rescaling and padding for the segmentation model.
For audio inputs, we follow settings in \cite{zhou2022avs} to use the 5-second audio clips and convert to 3 fixed-sized embeddings after padding, since the ImageBind \cite{girdhar2023imagebind} audio encoder samples 2-second audio each time.

To ensure training efficiency and preserve generalization ability, we freeze the vision encoders and audio encoder. 
Fine-tuning of the LLM is conducted using LoRA \cite{hu2021lora}, and the trainable parameters comprise the mask decoder and projection layers, accounting for approximately 7\% of the total parameters.

% We adopt 8 NVIDIA V100 GPUs for training, with a batch size of 6 on each GPU, and set the gradient accumulation step to 8.
We conduct training using 8 NVIDIA V100 GPUs, each with a batch size of 6, and employ a gradient accumulation step set to 8.
% Mixed precision training is implemented, converting the vision and audio encoder to float16 precision.
The training utilizes mixed precision, converting both the vision and audio encoder to float16 precision.
% The optimizer is AdamW \cite{loshchilov2017adamw} with a learning rate of 5e-5 and weight decay of 0.01.
AdamW \cite{loshchilov2017adamw} optimizer with a learning rate of 5e-5 and weight decay of 0.01 is employed, alongside a cosine annealing scheduler incorporating 200 warmup steps.
% Additionally, a cosine annealing scheduler with 200 warmup steps is utilized.
% For LoRA, we select a rank of 8, alpha of 16, and apply LoRA exclusively on query and value projections in the LLM.
LoRA operates with the rank of 8 and alpha of 16, exclusively applied to query and value projections within the LLM.
We employ ZeRO stage-2 \cite{rajbhandari2020zero} with DeepSpeed \cite{rasley2020deepspeed} which completes network training in 10K steps.
%and run for 10k steps which takes 3-4 days.
% We utilize ZeRO stage-2 \cite{rajbhandari2020zero} with DeepSpeed \cite{rasley2020deepspeed}, and train for 10k steps.

\begin{table*}[t]
\centering
\resizebox{0.75\textwidth}{!}
{
\begin{tabular}{llccclccclcc}
\toprule
\multirow{2}{*}{Method} &  & \multicolumn{3}{c}{RefCOCO} &  & \multicolumn{3}{c}{RefCOCO+} &  & \multicolumn{2}{c}{RefCOCOg} \\
% \cmidrule{3-12} 
\cline{3-12}
 &  & val & testA & testB &  & val & testA & testB &  & val(U) & test(U) \\ 
% \midrule
\hline
\multicolumn{12}{c}{\emph{\textbf{\color{gray}Specialist Segmentation Models}}} \\
CRIS \cite{wang2022cris} &  & 70.5 & 73.2 & 66.1 &  & 65.3 & 68.1 & 53.7 &  & 59.9 & 60.4 \\
LAVT \cite{yang2022lavt} &  & 72.7 & 75.8 & 68.8 &  & 62.1 & 68.4 & 55.1 &  & 61.2 & 62.1 \\
GRES \cite{liu2023gres} &  & 73.8 & 76.5 & 70.2 &  & 66.0 & 71.0 & 57.7 & & 65.0 & 66.0 \\
PolyFormer \cite{liu2023polyformer} &  & 76.0 & 78.3 & 73.3 &  & 69.3 & \underline{74.6} & \underline{61.9} & & 69.2 & 70.2 \\
UNINEXT \cite{UNINEXT} &  & \bf{82.2} & \bf{83.4} & \bf{81.3} &  & \bf{72.5} & \bf{76.4} & \bf{66.2} & & \bf{74.7} & \bf{76.4} \\
SEEM \cite{zou2023seem} &  & - & - & - &  & - & - & - & & 65.7 & - \\
\hline
% \emph{\textbf{Generalist MLLMs}} &  & \multicolumn{1}{l}{} & \multicolumn{1}{l}{} & \multicolumn{1}{l}{} &  & \multicolumn{1}{l}{} & \multicolumn{1}{l}{} & \multicolumn{1}{l}{} &  & \multicolumn{1}{l}{} & \multicolumn{1}{l}{} \\
\multicolumn{12}{c}{\emph{\textbf{\color{gray}Generalist MLLMs}}} \\
X-Decoder \cite{zou2023xdecoder} &  & - & - & - &  & - & - & - &  & 64.6 & - \\
LISA-7B \cite{lai2023lisa} &  & 74.1 & 76.5 & 71.1 &  & 62.4 & 67.4 & 56.5 &  & 66.4 & 68.4 \\
LISA-7B (ft) \cite{lai2023lisa} &  & 74.9 & 79.1 & 72.3 &  & 65.1 & 70.8 & 58.1 &  & 67.9 & 70.6 \\
\rowcolor{orange!10}
{\bf{AnyRef}} &  & 74.1 & 75.5 & 70.8 &  & 64.1 & 68.7 & 57.5 &  & 68.1 & 69.9 \\
\rowcolor{orange!10}
{\bf{AnyRef}} (ft) &  & \underline{76.9} & \underline{79.9} & \underline{74.2} &  & \underline{70.3} & {73.5} & 61.8 &  & \underline{70.0} & \underline{70.7} \\
\bottomrule
\end{tabular}
}
\caption{{\bf{Referring expression segmentation}} results (cIOU) on RefCOCO(+/g) datasets. (ft) denotes finetuning the model on RefCOCO(+/g) datasets.
Our model surpasses all generalist models and most specialist (segmentation-oriented) models.}
\label{res}
\vspace{-0.3cm}
\end{table*}

% {\textbf{Datasets.}}
\subsubsection{Datasets}
\label{methoddatasets}
The training process involves a diverse range of datasets.
For general semantic and instance segmentation, COCO-Stuff \cite{caesar2018cocostuff}, ADE20K \cite{zhou2019ade20k}, and PACO-LVIS \cite{ramanathan2023paco} are utilized, with one category chosen per batch. 
Referring expression segmentation incorporates RefClef, RefCOCO, RefCOCO+ \cite{kazemzadeh2014referitgame}, RefCOCOg \cite{yu2016refcocog}, and PhraseCut \cite{wu2020phrasecut}.
% Image-level referring segmentation follows the approach in \cite{luddecke2022imagetextprompt}, selecting samples from COCO \cite{lin2014coco}, PascalVOC \cite{everingham2010pascal} and PhraseCut \cite{wu2020phrasecut} with a random cropped draw from images containing the same category with its corresponding linguist expression.
Image-level referring segmentation adopts the method outlined in \cite{luddecke2022imagetextprompt}, where samples are chosen from COCO \cite{lin2014coco}, PascalVOC \cite{everingham2010pascal}, and PhraseCut \cite{wu2020phrasecut} datasets. Random cropped samples are drawn from images that contain the same category as their corresponding linguistic expressions.
Region-level captioning involves RefCOCO(+/g) and Flickr30K Entities \cite{flickrentities}.
Audio-visual segmentation employs AVSBench \cite{zhou2022avs} with both single and multiple sound sources.
% Additionally, LLaVA-Instruct-150k \cite{liu2023llava} is included to preserve the reasoning ability of the MLLM.
To prevent data leakage, samples with images in the validation or test splits are excluded.

\section{Experiments}

We assess the capabilities of our model through evaluations on various benchmarks, including different modality referring segmentation (text/image/audio) for pixel-level perception and referring expression generation for regional understanding.
Models are categorized as \emph{specialists} or \emph{generalists}, with the former designed exclusively for specific tasks.
We provide examples for each task in \cref{fig:task}, and more illustrations can be found in the supplementary material.

% \begin{table}[htbp]
% \resizebox{\columnwidth}{!}
% {
% \begin{tabular}{lcccc}
% % \hline
% \toprule
% \multirow{2}{*}{Method} & \multicolumn{2}{c}{$\text{COCO-20}^i$} & \multicolumn{2}{c}{$\text{RASCAL-5}^i$} \\
% \cline{2-5} 
%  & one-shot & few-shot & one-shot & few-shot \\
% \hline
% % \midrule
% \textit{specialist model} & \multicolumn{1}{l}{} & \multicolumn{1}{l}{} & \multicolumn{1}{l}{} & \multicolumn{1}{l}{} \\
% HSNet* \cite{min2021hsnet} & 41.7 & 50.7 & 68.7 & 73.8 \\
% VAT* \cite{hong2022vat} & 42.9 & 49.4 & 72.4 & 76.3 \\
% % FPTrans* \cite{zhang2022fptrans} & \bf{56.5} & \underline{65.5} & \underline{77.7} & \underline{83.2} \\
% CLIPSeg \cite{lueddecke22clipseg} & 33.2 & - & 59.5 & - \\
% SegGPT \cite{wang2023seggpt} & \bf{56.1} & \bf{67.9} & \bf{83.2} & \bf{89.8} \\
% \hline
% \textit{generalist model} & \multicolumn{1}{l}{} & \multicolumn{1}{l}{} & \multicolumn{1}{l}{} & \multicolumn{1}{l}{} \\
% Painter \cite{wang2023painter} & 32.8 & 32.6 & 64.5 & 64.6 \\

% \rowcolor{orange!10}
% {\bf{AnyRef}} & 43.5 & 51.3 & 74.8 & 78.6 \\
% \rowcolor{orange!10}
% {\bf{AnyRef}}$\dagger$ & \underline{46.3} & \underline{55.2} & \underline{76.5} & \underline{80.0} \\
% % \hline
% \bottomrule
% \end{tabular}
% }
% \caption{Quantitative results of {\bf{example-based few-shot segmentation}}. 
% % * indicates in-domain settings as in \cite{wang2023seggpt}.
% * indicates that the categories in training cover that in testing as in \cite{wang2023seggpt}, and $\dagger$ denotes using mask cropping setting.
% % segmentation masks along with reference images.}
% }
% \label{fewshot}
% \end{table}

\subsection{Multi-modality Referring Segmentation}

\subsubsection{Referring Expression Segmentation}
% \noindent{\bf{Referring Expression Segmentation.}}
The task involves labeling pixels within an image corresponding to an object instance referred to by a linguistic expression.
We instruct our model as: ``\texttt{Can you segment \{exp\} in this image?}", where \texttt{\{exp\}} is the given explicit description.
Evaluation is conducted using Cumulative-IoU (cIoU) as the metric.
We make comparisons with state-of-the-art models on validation and test sets of RefCOCO, RefCOCO+ and RefCOCOg \cite{kazemzadeh2014referitgame, yu2016refcocog}.
As shown in \cref{res}, our performance surpasses all generalist models and most specialist models except UNINEXT-H \cite{UNINEXT}, which is trained using a considerably larger dataset that includes video samples.
% The specialist models only have the ability of segmentation-related tasks, while generalist models have additional capacity of output textural descriptions.
Specialist models excel solely at segmentation-related tasks, while generalist models possess additional capabilities for generating textural descriptions and are capable of handling more complex references.

\subsubsection{Image Referring Segmentation}
Predicting masks using image examples is akin to one- or few-shot segmentation, where regions corresponding to the highlighted object in the example image must be located in a query image.
We prompt our model with queries like ``\texttt{Can you find similar object of {\textless{\texttt{img\_ref}}\textgreater}{\textless{\texttt{img\_feat}}\textgreater}{\textless{\texttt{/img\_ref}}\textgreater} in this image?}", where {\textless{\texttt{img\_feat}}\textgreater} denotes pooled features from example images as detailed in \cref{unifiedreferringrepresentation}.
The evaluation takes place under the in-domain setting on $\text{COCO-20}^i$ \cite{lin2014coco} and $\text{PASCAL-5}^i$ \cite{everingham2010pascal} for a fair comparison, as most classes are encountered during the training stages.
In the few-shot evaluation, the model inferences multiple times using different example images, with the averaged mask serving as the final prediction.
In our referring examples, we do not have corresponding mask examples, which is different from the standard setting.
we follow \cite{lueddecke22clipseg} to crop out the target object for highlighting, using their segmentation masks.
As demonstrated in \cref{fewshot}, our model achieves competitive performance compared to state-of-the-art methods.
% Notably, when using the referring mask cropping setting, the performance gains are enhanced as the reference provides clearer instructions for identifying the target object.

\begin{table}[tbp]
\resizebox{\columnwidth}{!}
{
\begin{tabular}{lcccc}
% \hline
\toprule
\multirow{2}{*}{Method} & \multicolumn{2}{c}{$\text{COCO-20}^i$} & \multicolumn{2}{c}{$\text{RASCAL-5}^i$} \\
\cline{2-5} 
 & one-shot & few-shot & one-shot & few-shot \\
\hline
% \midrule
% \textit{\textbf{Specialist Segmentation Models}} & \multicolumn{1}{l}{} & \multicolumn{1}{l}{} & \multicolumn{1}{l}{} & \multicolumn{1}{l}{} \\
\multicolumn{5}{c}{\emph{\textbf{\color{gray}Specialist Segmentation Models}}} \\
HSNet* \cite{min2021hsnet} & 41.7 & 50.7 & 68.7 & 73.8 \\
VAT* \cite{hong2022vat} & 42.9 & 49.4 & 72.4 & 76.3 \\
% FPTrans* \cite{zhang2022fptrans} & \bf{56.5} & \underline{65.5} & \underline{77.7} & \underline{83.2} \\
CLIPSeg \cite{lueddecke22clipseg} & 33.2 & - & 59.5 & - \\
SegGPT \cite{wang2023seggpt} & \bf{56.1} & \bf{67.9} & \bf{83.2} & \bf{89.8} \\
\hline
% \textit{\textbf{Generalist MLLMs}} & \multicolumn{1}{l}{} & \multicolumn{1}{l}{} & \multicolumn{1}{l}{} & \multicolumn{1}{l}{} \\
\multicolumn{5}{c}{\emph{\textbf{\color{gray}Generalist Multi-task Models}}} \\
Painter \cite{wang2023painter} & 32.8 & 32.6 & 64.5 & 64.6 \\

\rowcolor{orange!10}
{\bf{AnyRef}} & 43.5 & 51.3 & 74.8 & 78.6 \\
\rowcolor{orange!10}
{\bf{AnyRef}}$\dagger$ & \underline{46.3} & \underline{55.2} & \underline{76.5} & \underline{80.0} \\
% \hline
\bottomrule
\end{tabular}
}
\caption{Quantitative results of {\bf{example-based few-shot segmentation}}. 
% * indicates in-domain settings as in \cite{wang2023seggpt}.
* indicates that the categories in training cover that in testing as in \cite{wang2023seggpt}, and $\dagger$ denotes using mask cropping setting.
% segmentation masks along with reference images.}
}
\label{fewshot}
\end{table}

\subsubsection{Audio-Visual Segmentation}
The AVS benchmark comprises single- and multi-sources subsets based on the number of sounding objects.
We utilize prompts like, ``\texttt{Can you segment the object(s) that produce sound of {\textless{\texttt{aud\_ref}}\textgreater}{\textless{\texttt{aud\_feat}}\textgreater}{\textless{\texttt{/aud\_ref}}\textgreater} in this image?}", to instruct the model for mask predictions.
Following \cite{zhou2022avs}, evaluation metrics include mean IoU (mIoU) for region similarity and F-score\footnote{$F_\beta = \frac{(1+\beta^2) \times \text{precision} \times \text{recall}}{\beta^2 \times \text{precision} + \text{recall}}$, where $\beta^2 =0.3$ following \cite{zhou2022avs}} for contour accuracy.
The quantitative results in \cref{avs} demonstrate that our model consistently outperforms most methods on single-source split, indicating successful alignment of audio features with the LLM during fine-tuning.
However, when confronted with audios containing multiple sound sources, our model encounters challenges in producing masks that cover more than one object.
Moreover, owing to the ability of LLM, our model can determine the textural category of the sounding objects, as depicted in \cref{fig:task} (d).

\begin{table}[tbp]
\centering
\resizebox{0.9\columnwidth}{!}
{
\begin{tabular}{lccccc}
\toprule
\multirow{2}{*}{Method} & \multicolumn{2}{c}{Single-source} & & \multicolumn{2}{c}{Multi-source} \\
\cline{2-6} 
% \cmidrule{2-5}
 & mIOU & F-score & & mIOU & F-score \\
\hline
% \midrule
AVS \cite{zhou2022avs} & 78.7 & 0.879 & & 54.0 & 0.645 \\
BG \cite{hao2023avsbg} & 81.7 & 0.904 & & 55.1 & 0.668 \\
AVSegformer \cite{gao2023avsegformer} & 82.1 & 0.899 & & \underline{58.4} & \underline{0.693} \\
AUSS \cite{ling2023heartoseg} & \bf{89.4} & \bf{0.942} & & \bf{63.5} & \bf{0.752} \\
\midrule
% \textit{generalist model} & \multicolumn{1}{l}{} & \multicolumn{1}{l}{} & \multicolumn{1}{l}{} & \multicolumn{1}{l}{} \\
\rowcolor{orange!10}
{\bf{AnyRef}} & \underline{82.8} & \underline{0.908} & & 55.6 & 0.663 \\
% \hline
\bottomrule
\end{tabular}
}
\caption{Quantitative results of {\bf{audio-visual segmentation}}.}
\label{avs}
\vspace{-0.3cm}
\end{table}

\subsection{Referring Expression Generation}
\label{refexpgen}
This task involves generating a textual description associated with an object based on its location (bounding box).
We evaluate our generated expressions using automatic caption generation metrics, including CIDEr \cite{vedantam2015cider} and Meteor \cite{lavie-agarwal-2007-meteor}, on RefCOCO, RefCOCO+ and RefCOCOg.
Our model achieves remarkable performance among generalist LLM-based models and demonstrates competitive result to specialist models, as shown in \cref{table:reg}.

\begin{figure}[t]
    \centering
    \includegraphics[width=\columnwidth]{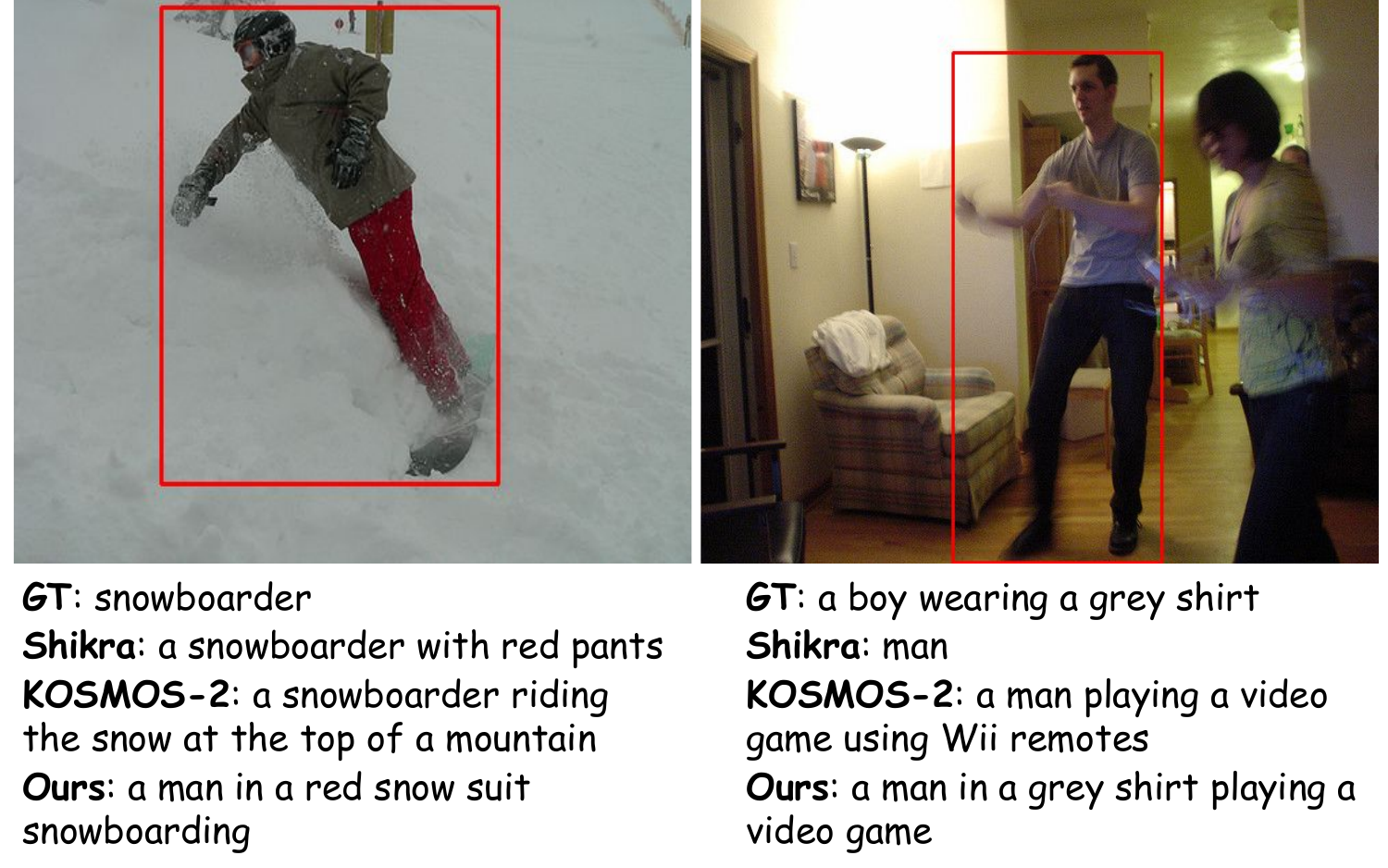}
    \caption{Comparison of {\bf{generated expressions}} between ground-truth and LLM-based methods.}
    % The latter generates more detailed descriptions.}
    \label{reg}
    \vspace{-0.4cm}
\end{figure}

Nonetheless, as stated in \cite{yu2017slr, liu2017attribute, bracha2023disclip}, standard automated evaluation metrics do not authentically capture generation quality due to the constraints of ground-truth expressions.
This scenario is particularly pronounced in open-text generation, especially for LLM-based models.
These models have the ability to generate rich, natural sentences, while the provided ground-truth expressions often tend to be concise, as indicated in \cref{reg}.
% , as they generate rich natural sentences while the ground-truth expressions tend to be terse, as indicated in \cref{reg}.

% \begin{figure}[htbp]
%     \centering
%     \includegraphics[width=0.7\columnwidth]{cluster.pdf}
%     \caption{{\bf{Visualization of mask embeddings}} before and after the \emph{refocusing mechanism}. {\bf{\color{black} original}} denotes original mask embeddings of all categories, while {\bf{\color{red}vehicle}}, {\bf{\color{blue}person}}, and {\bf{\color{OliveGreen}animal}} represent the updated mask embeddings of different categories.
%     }
%     \label{fig:cluster}
% \end{figure}

To further evaluate the quality of the generated expressions, we conduct human evaluations following \cite{yu2016modeling, yu2017slr, bracha2023disclip}.
We randomly select 100 images from the validation datasets and ask five human raters to choose the bounding box that best matches the generated expression, and the averaged score is considered the final result.
In \cref{human}, we present the results of the human evaluations, including both traditional methods and LLM-based methods.
The LLM-based methods produce more detailed descriptions, closely resembling human behavior, which are preferred by the human raters.
We provide more examples in supplementary material.

\begin{table*}[htb]
\centering
\resizebox{0.9\textwidth}{!}
{
\begin{tabular}{llcccclcccclcc}
\toprule
\multirow{3}{*}{Method} &  & \multicolumn{4}{c}{RefCOCO} &  & \multicolumn{4}{c}{RefCOCO+} &  & \multicolumn{2}{c}{RefCOCOg} \\
\cline{3-6} \cline{8-11} \cline{13-14} 
% \cmidrule{3-6} \cmidrule{8-11} \cmidrule{13-14} 
 &  & \multicolumn{2}{c}{testA} & \multicolumn{2}{c}{testB} &  & \multicolumn{2}{c}{testA} & \multicolumn{2}{c}{testB} &  & \multicolumn{2}{c}{val} \\
 \cline{3-6} \cline{8-11} \cline{13-14} 
 % \cmidrule{3-6} \cmidrule{8-11} \cmidrule{13-14} 
 &  & Meteor & CIDEr & Meteor & CIDEr &  & Meteor & CIDEr & Meteor & CIDEr &  & Meteor & CIDEr \\
% \midrule
\hline
\emph{\textbf{Specialist Models}} &  & \multicolumn{1}{l}{} & \multicolumn{1}{l}{} & \multicolumn{1}{l}{} & \multicolumn{1}{l}{} &  & \multicolumn{1}{l}{} & \multicolumn{1}{l}{} & \multicolumn{1}{l}{} & \multicolumn{1}{l}{} &  & \multicolumn{1}{l}{} & \multicolumn{1}{l}{} \\
Visdif \cite{yu2016visdif} & & 18.5 & - & 24.7 & - & & 14.2 & - & 13.5 & - & & 14.5 & - \\
SLR \cite{yu2017slr} & & 29.6 & 77.5 & 34.0 & 132.0 & & 21.3 & 52.0 & 21.5 & 73.5 & & 15.9 & 66.2 \\
easyREG \cite{tanaka2019easytounderstand} & & \underline{31.3} & \underline{83.7} & \underline{34.1} & 132.9 & & \underline{24.2} & 66.4 & 22.8 & 78.7 & & 17.0 & 77.7 \\
IREG \cite{ye2023ireg} &  & \bf{34.9} & \bf{105.4} & \bf{37.3} & \bf{154.1} &  & \bf{30.8} & \bf{89.8} & \bf{26.4} & \bf{97.0} &  & \bf{19.4} & \bf{101.2} \\ \hline
\emph{\textbf{Generalist MLLMs}} &  & \multicolumn{1}{l}{} & \multicolumn{1}{l}{} & \multicolumn{1}{l}{} & \multicolumn{1}{l}{} &  & \multicolumn{1}{l}{} & \multicolumn{1}{l}{} & \multicolumn{1}{l}{} & \multicolumn{1}{l}{} &  & \multicolumn{1}{l}{} & \multicolumn{1}{l}{} \\
GRIT \cite{wu2022grit} &  & - & - & - & - &  & - & - & - & - &  & 15.2 & 71.6 \\
KOSMOS-2 \cite{peng2023kosmos}  &  & - & - & - & - &  & - & - & - & - &  & 14.1 & 62.3 \\
% Shikra \cite{chen2023shikra} &  & & & & &  & & & & &  & & \\
\rowcolor{orange!10} 
{\bf{AnyRef}} & & 23.9 & 74.8 & 26.7 & 118.6 & & 16.4 & 59.4 & 14.3 & 62.9 & & 16.2 & 69.0 \\
\rowcolor{orange!10}
{\bf{AnyRef}} (ft) &  & 30.4 & 79.5 & 32.7 & \underline{138.6} &  & 23.2 & \underline{67.7} & 20.1 & \underline{80.1} &  & \underline{17.1} & \underline{79.7} \\
\bottomrule
\end{tabular}
}
\caption{Quantitative results on {\bf{region-level referring expression generation}}. \emph{Generalist models} (LLM-based) perform poorly on automated evaluation metrics due to the limitation of constrained ground-truth expressions, as stated in \cref{refexpgen}.}
\label{table:reg}
\vspace{-0.3cm}
\end{table*}

\begin{table}[tbp]
\centering
\resizebox{0.9\columnwidth}{!}
{
\begin{tabular}{lcccccc}
\toprule
\multirow{2}{*}{Method} & & \multicolumn{2}{c}{RefCOCO} & & \multicolumn{2}{c}{RefCOCO+}  \\
\cline{3-7}
% \cmidrule{2-5}
& & testA & testB & & testA & testB  \\
\hline
% \midrule
SLR \cite{yu2017slr} & & 66\% & 62\% & & 43\% & 38\% \\
SLR+Rerank \cite{yu2017slr} & & 73\% & 77\% & & 49\% & 46\% \\
% \hline
\midrule
KOSMOS-2 \cite{peng2023kosmos} & & 88\% & \bf{84}\% & & 63\% & 65\%  \\
Shikra \cite{chen2023shikra} & & \bf{91}\% & 81\% & & 59\% & 62\% \\
\rowcolor{orange!10}
{\bf{AnyRef}} & & 87\% & 80\% & & \bf{67\%} & \bf{66\%} \\
% \hline
\bottomrule
\end{tabular}
}
\caption{{\bf{Human evaluation}} on referring expression generation.}
\label{human}
\end{table}

\subsection{Ablation Study}

\begin{figure}[bp]
    \centering
    \includegraphics[width=0.8\columnwidth]{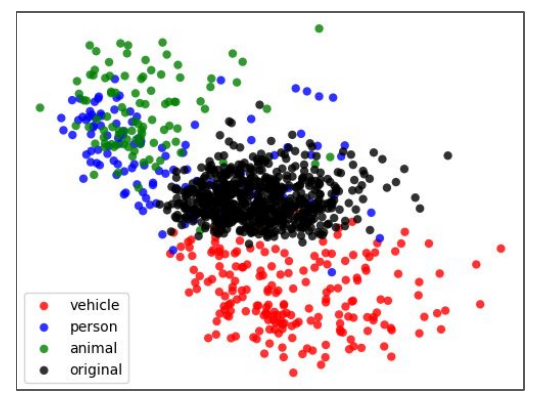}
    \caption{Visualization of mask embeddings before and after the \emph{refocusing mechanism}. {\bf{\color{black} original}} denotes original mask embeddings, while {\bf{\color{red}vehicle}}, {\bf{\color{blue}person}}, and {\bf{\color{OliveGreen}animal}} represent the updated mask embeddings corresponding to their respective referring objects contained in the textural expression.}
    \label{fig:cluster}
    \vspace{-0.3cm}
\end{figure}

We conduct extensive ablation studies to reveal the contribution of each component.

\begin{table}[tbp]
\centering
\resizebox{0.9\columnwidth}{!}
{
\begin{tabular}{l|cccccc}
\toprule
\multirow{2}{*}{$\lambda_f$} & RefCOCOg  &  & AVSBench &  & \multicolumn{2}{c}{RefCOCOg} \\ \cline{2-2} \cline{4-4} \cline{6-7} 
 & cIOU     &  & mIOU     &  & Meteor        & CIDEr \\
\midrule
0.0 & 68.7 & & 81.4 & & 16.8 & 71.1 \\
1.0 & 67.1 & & 80.6 & & 14.3 & 68.5 \\
\rowcolor{orange!10}
0.1 & \bf{70.0} &  & \bf{82.8} &  & \bf{17.1} & 73.7 \\
\midrule
1.0$\dagger$ & 68.0 &  & 81.1 &  & 15.7 & 70.0 \\
0.1$\dagger$ & 69.3 &  & 82.0 &  & 17.0 & \bf{73.8} \\ 
\bottomrule
\end{tabular}
}
\caption{Ablation study on {\bf{refocusing weight $\lambda_f$}}. $x\dagger$ indicates trainable $\lambda_f$ initialized with $x$.}
\label{lambdaf}
\vspace{-0.3cm}
\end{table}

% \begin{figure}[bp]
%     \centering
%     \includegraphics[width=0.7\columnwidth]{cluster.pdf}
%     \caption{{\bf{Visualization of mask embeddings}} before and after the \emph{refocusing mechanism}. {\bf{\color{black} original}} denotes original mask embeddings of all categories, while {\bf{\color{red}vehicle}}, {\bf{\color{blue}person}}, and {\bf{\color{OliveGreen}animal}} represent the updated mask embeddings of different categories.
%     %of corresponding to their respective referring objects contained in the textural expression.
%     }
%     \label{fig:cluster}
% \end{figure}

\noindent{\bf{Refocusing Mechanism.}}
We first investigate the effectiveness of enhancing the {\textless{\texttt{obj}}\textgreater} token through \emph{refocusing mechanism}, and explore the impact of different refocusing weights $\lambda_f$.
We evaluate setting different values for $\lambda_f$ and also try setting it as a learnable parameter along with the model.
We conduct evaluations on both referring segmentation and expression generation tasks.
Results in \cref{lambdaf} reveal that the refocusing weight significantly affects performance in both tasks.
A small weight of 0.1 improves performance, while a larger weight can have detrimental effects, particularly in expression generation.
We also experiment with learning $\lambda_f$ as a parameter along with the model, but we find that the performance varies greatly depending on the initialized value.
Thus, for simplicity and stability, we empirically select $\lambda_f = 0.1$ for our experiments.

We further employ PCA to visualize the mask embeddings before and after implementing the \emph{refocusing mechanism} in \cref{fig:cluster}
We choose three subsets representing different referring objects including vehicles, persons and animals (\emph{e.g.}, the person subset comprises output expressions containing ``person," ``man," ``woman," etc.).
% The visualization reveals a broader representation range in the mask embedding after applying the \emph{refocusing mechanism}, and the updated embeddings are clustered according to the textual expressions, aiding in more accurate decoding of masks.
% It shows that with the representation range of the mask embedding is wider after the \emph{refocusing mechanism}, and embeddings are clustered according to the textural expression, which advantages for decoding the correct mask.
The visualization illustrates that the \emph{refocusing mechanism} results in a wider representation range of the mask embedding.
Moreover, the updated embeddings demonstrate a clustering pattern aligned with the associated textual expressions, contributing to a more precise decoding of masks.

% \noindent{\bf{Unified Referring Representation.}}
% We also evaluate the effect of different number of referring representations.

% number of tokens? 
% current 4x4
% kosmos 4
% gpt4roi 14*14

\noindent{\bf{Training Datasets.}}
The impact of different types of datasets is validated in \cref{ablationdatasets}, and evaluation is carried out on RefCOCOg validation split.
Region/Image Ref. refers to region-level and image-level referring data, as explained in \cref{methoddatasets}.
It becomes apparent that the model's generalization improves as the type of datasets increases.

\newcommand{\blackcmark}{\textcolor{black}{\ding{51}}}

\begin{table}[tbp]
\centering
\resizebox{0.9\columnwidth}{!}
{
\begin{tabular}{c|cccc|c}
\toprule
Exp. & Referring & General & Region Ref. & Image Ref. & cIOU \\
\midrule
1    & \blackcmark &         &            &              & 66.2 \\
2    & \blackcmark & \blackcmark &            &              & 67.0 \\
3    & \blackcmark & \blackcmark & \blackcmark &              & 67.7 \\
4    & \blackcmark & \blackcmark &            & \blackcmark & 67.4 \\
5    & \blackcmark & \blackcmark & \blackcmark & \blackcmark & \bf{68.1} \\
\bottomrule
\end{tabular}
}
\caption{Ablation study on {\bf{training datasets}}.}
\label{ablationdatasets}
\vspace{-0.3cm}
\end{table}

% \noindent{\bf{Mixing Multi-modality Referring.}}
% We further investigate whether using multiple references from different modalities improves performances, \emph{e.g.}, giving textural description and audio inputs simultaneously.
% However, we obverse no performance improvement.
% We conjecture that with the powerful understanding and reasoning abilities of the LLM and multi-modal feature alignments, the LLM can transfer different modalities into the textural space, therefore prompting the same concepts multiple times will not boost its performance but potentially confuse the LLM.
% However, this is still an explored area which needs further investigation, which beyond the scope of this paper, and we provide more details in the supplementary.

\section{Conclusion}

We present {\bf{AnyRef}}, a pioneering MLLM model capable of generating pixel-level object perceptions and language descriptions from various modality references, including texts, regions, images, and audio. 
% This is achieved through the unified referring representation, which provides an interface between different modality inputs and the LLM. 
This is made possible by the unified referring representation, which connects different types of inputs to the LLM.
% We further propose a refocusing mechanism which can aggregate the grounding expressions of referenced objects through attention scores to enhance the segmentation embedding, giving rise to superior pixel-level vision perception ability.
We further propose a refocusing mechanism that uses attention scores to improve the segmentation embedding and enhance pixel-level vision perception.
Across various downstream tasks, our model exhibits remarkable performance while providing users with enhanced interacting flexibility.
%
%
%The proposed refocusing mechanism prioritizes the referenced object and implicitly incorporates pixel-level supervision.
%It efficiently utilizes attention scores from LLM inference without requiring additional computations.
% Our model demonstrates remarkable performance on multiple downstream tasks, including multi-modality referring segmentation and referring expression generation, while providing users with enhanced flexibility in interacting with the model.
% Across various downstream tasks such as multi-modality referring segmentation and referring expression generation, our model exhibits remarkable performance, while providing users with enhanced interacting flexibility.

\noindent \textbf{Acknowledgements.}
This work is supported by the National Natural Science Foundation of China (U23A20386, 62276045, 62293540, 62293542), Dalian Science and Technology Talent Innovation Support Plan (2022RY17).

\clearpage
{
    \small
    \bibliographystyle{ieeenat_fullname}
    \bibliography{main}

\begin{thebibliography}{63}
\providecommand{\natexlab}[1]{#1}
\providecommand{\url}[1]{\texttt{#1}}
\expandafter\ifx\csname urlstyle\endcsname\relax
  \providecommand{\doi}[1]{doi: #1}\else
  \providecommand{\doi}{doi: \begingroup \urlstyle{rm}\Url}\fi

\bibitem[Alayrac et~al.(2022)Alayrac, Donahue, Luc, Miech, Barr, Hasson, Lenc, Mensch, Millican, Reynolds, et~al.]{alayrac2022flamingo}
Jean-Baptiste Alayrac, Jeff Donahue, Pauline Luc, Antoine Miech, Iain Barr, Yana Hasson, Karel Lenc, Arthur Mensch, Katherine Millican, Malcolm Reynolds, et~al.
\newblock Flamingo: a visual language model for few-shot learning.
\newblock \emph{Advances in Neural Information Processing Systems}, 35:\penalty0 23716--23736, 2022.

\bibitem[Bracha et~al.(2023)Bracha, Shaar, Shamsian, Fetaya, and Chechik]{bracha2023disclip}
Lior Bracha, Eitan Shaar, Aviv Shamsian, Ethan Fetaya, and Gal Chechik.
\newblock Disclip: Open-vocabulary referring expression generation.
\newblock \emph{arXiv preprint arXiv:2305.19108}, 2023.

\bibitem[Caesar et~al.(2018)Caesar, Uijlings, and Ferrari]{caesar2018cocostuff}
Holger Caesar, Jasper Uijlings, and Vittorio Ferrari.
\newblock Coco-stuff: Thing and stuff classes in context.
\newblock In \emph{Proceedings of the IEEE conference on computer vision and pattern recognition}, pages 1209--1218, 2018.

\bibitem[Chen et~al.(2023)Chen, Zhang, Zeng, Zhang, Zhu, and Zhao]{chen2023shikra}
Keqin Chen, Zhao Zhang, Weili Zeng, Richong Zhang, Feng Zhu, and Rui Zhao.
\newblock Shikra: Unleashing multimodal llm's referential dialogue magic.
\newblock \emph{arXiv preprint arXiv:2306.15195}, 2023.

\bibitem[Chen et~al.(2022)Chen, Saxena, Li, Lin, Fleet, and Hinton]{chen2022pix2seq2}
Ting Chen, Saurabh Saxena, Lala Li, Tsung-Yi Lin, David~J Fleet, and Geoffrey~E Hinton.
\newblock A unified sequence interface for vision tasks.
\newblock \emph{Advances in Neural Information Processing Systems}, 35:\penalty0 31333--31346, 2022.

\bibitem[Dai et~al.(2023)Dai, Li, Li, Tiong, Zhao, Wang, Li, Fung, and Hoi]{instructblip}
Wenliang Dai, Junnan Li, Dongxu Li, Anthony Meng~Huat Tiong, Junqi Zhao, Weisheng Wang, Boyang Li, Pascale Fung, and Steven Hoi.
\newblock Instructblip: Towards general-purpose vision-language models with instruction tuning, 2023.

\bibitem[Everingham et~al.(2010)Everingham, Van~Gool, Williams, Winn, and Zisserman]{everingham2010pascal}
Mark Everingham, Luc Van~Gool, Christopher~KI Williams, John Winn, and Andrew Zisserman.
\newblock The pascal visual object classes (voc) challenge.
\newblock \emph{International journal of computer vision}, 88:\penalty0 303--338, 2010.

\bibitem[Gao et~al.(2023)Gao, Chen, Chen, Wang, and Lu]{gao2023avsegformer}
Shengyi Gao, Zhe Chen, Guo Chen, Wenhai Wang, and Tong Lu.
\newblock Avsegformer: Audio-visual segmentation with transformer.
\newblock \emph{arXiv preprint arXiv:2307.01146}, 2023.

\bibitem[Girdhar et~al.(2023)Girdhar, El-Nouby, Liu, Singh, Alwala, Joulin, and Misra]{girdhar2023imagebind}
Rohit Girdhar, Alaaeldin El-Nouby, Zhuang Liu, Mannat Singh, Kalyan~Vasudev Alwala, Armand Joulin, and Ishan Misra.
\newblock Imagebind: One embedding space to bind them all.
\newblock In \emph{CVPR}, 2023.

\bibitem[Hao et~al.(2023)Hao, Mao, He, Han, Dai, and Zhong]{hao2023avsbg}
Dawei Hao, Yuxin Mao, Bowen He, Xiaodong Han, Yuchao Dai, and Yiran Zhong.
\newblock Improving audio-visual segmentation with bidirectional generation.
\newblock \emph{arXiv preprint arXiv:2308.08288}, 2023.

\bibitem[He et~al.(2017)He, Gkioxari, Doll{\'a}r, and Girshick]{he2017maskrcnn}
Kaiming He, Georgia Gkioxari, Piotr Doll{\'a}r, and Ross Girshick.
\newblock Mask r-cnn.
\newblock In \emph{Proceedings of the IEEE international conference on computer vision}, pages 2961--2969, 2017.

\bibitem[Hong et~al.(2022)Hong, Cho, Nam, Lin, and Kim]{hong2022vat}
Sunghwan Hong, Seokju Cho, Jisu Nam, Stephen Lin, and Seungryong Kim.
\newblock Cost aggregation with 4d convolutional swin transformer for few-shot segmentation.
\newblock In \emph{European Conference on Computer Vision}, pages 108--126. Springer, 2022.

\bibitem[Hu et~al.(2021)Hu, Shen, Wallis, Allen-Zhu, Li, Wang, Wang, and Chen]{hu2021lora}
Edward~J Hu, Yelong Shen, Phillip Wallis, Zeyuan Allen-Zhu, Yuanzhi Li, Shean Wang, Lu Wang, and Weizhu Chen.
\newblock Lora: Low-rank adaptation of large language models.
\newblock \emph{arXiv preprint arXiv:2106.09685}, 2021.

\bibitem[Kazemzadeh et~al.(2014)Kazemzadeh, Ordonez, Matten, and Berg]{kazemzadeh2014referitgame}
Sahar Kazemzadeh, Vicente Ordonez, Mark Matten, and Tamara Berg.
\newblock Referitgame: Referring to objects in photographs of natural scenes.
\newblock In \emph{Proceedings of the 2014 conference on empirical methods in natural language processing (EMNLP)}, pages 787--798, 2014.

\bibitem[Kirillov et~al.(2023)Kirillov, Mintun, Ravi, Mao, Rolland, Gustafson, Xiao, Whitehead, Berg, Lo, Doll{\'a}r, and Girshick]{kirillov2023segany}
Alexander Kirillov, Eric Mintun, Nikhila Ravi, Hanzi Mao, Chloe Rolland, Laura Gustafson, Tete Xiao, Spencer Whitehead, Alexander~C. Berg, Wan-Yen Lo, Piotr Doll{\'a}r, and Ross Girshick.
\newblock Segment anything.
\newblock \emph{arXiv:2304.02643}, 2023.

\bibitem[Lai et~al.(2023)Lai, Tian, Chen, Li, Yuan, Liu, and Jia]{lai2023lisa}
Xin Lai, Zhuotao Tian, Yukang Chen, Yanwei Li, Yuhui Yuan, Shu Liu, and Jiaya Jia.
\newblock Lisa: Reasoning segmentation via large language model.
\newblock \emph{arXiv preprint arXiv:2308.00692}, 2023.

\bibitem[Lavie and Agarwal(2007)]{lavie-agarwal-2007-meteor}
Alon Lavie and Abhaya Agarwal.
\newblock {METEOR}: An automatic metric for {MT} evaluation with high levels of correlation with human judgments.
\newblock In \emph{Proceedings of the Second Workshop on Statistical Machine Translation}, pages 228--231, Prague, Czech Republic, 2007. Association for Computational Linguistics.

\bibitem[Li et~al.(2023{\natexlab{a}})Li, Zhang, Chen, Wang, Yang, and Liu]{li2023otter}
Bo Li, Yuanhan Zhang, Liangyu Chen, Jinghao Wang, Jingkang Yang, and Ziwei Liu.
\newblock Otter: A multi-modal model with in-context instruction tuning.
\newblock \emph{arXiv preprint arXiv:2305.03726}, 2023{\natexlab{a}}.

\bibitem[Li et~al.(2023{\natexlab{b}})Li, Li, Savarese, and Hoi]{li2023blip2}
Junnan Li, Dongxu Li, Silvio Savarese, and Steven Hoi.
\newblock {BLIP-2:} bootstrapping language-image pre-training with frozen image encoders and large language models.
\newblock In \emph{ICML}, 2023{\natexlab{b}}.

\bibitem[Lin et~al.(2014)Lin, Maire, Belongie, Hays, Perona, Ramanan, Doll{\'a}r, and Zitnick]{lin2014coco}
Tsung-Yi Lin, Michael Maire, Serge Belongie, James Hays, Pietro Perona, Deva Ramanan, Piotr Doll{\'a}r, and C~Lawrence Zitnick.
\newblock Microsoft coco: Common objects in context.
\newblock In \emph{Computer Vision--ECCV 2014: 13th European Conference, Zurich, Switzerland, September 6-12, 2014, Proceedings, Part V 13}, pages 740--755. Springer, 2014.

\bibitem[Ling et~al.(2023)Ling, Li, Gan, Zhang, Chi, and Wang]{ling2023heartoseg}
Yuhang Ling, Yuxi Li, Zhenye Gan, Jiangning Zhang, Mingmin Chi, and Yabiao Wang.
\newblock Hear to segment: Unmixing the audio to guide the semantic segmentation.
\newblock \emph{arXiv preprint arXiv:2305.07223}, 2023.

\bibitem[Liu et~al.(2023{\natexlab{a}})Liu, Ding, and Jiang]{liu2023gres}
Chang Liu, Henghui Ding, and Xudong Jiang.
\newblock Gres: Generalized referring expression segmentation.
\newblock In \emph{Proceedings of the IEEE/CVF Conference on Computer Vision and Pattern Recognition}, pages 23592--23601, 2023{\natexlab{a}}.

\bibitem[Liu et~al.(2023{\natexlab{b}})Liu, Li, Wu, and Lee]{liu2023llava}
Haotian Liu, Chunyuan Li, Qingyang Wu, and Yong~Jae Lee.
\newblock Visual instruction tuning.
\newblock In \emph{NeurIPS}, 2023{\natexlab{b}}.

\bibitem[Liu et~al.(2017)Liu, Wang, and Yang]{liu2017attribute}
Jingyu Liu, Liang Wang, and Ming-Hsuan Yang.
\newblock Referring expression generation and comprehension via attributes.
\newblock In \emph{Proceedings of the IEEE International Conference on Computer Vision}, pages 4856--4864, 2017.

\bibitem[Liu et~al.(2023{\natexlab{c}})Liu, Ding, Cai, Zhang, Satzoda, Mahadevan, and Manmatha]{liu2023polyformer}
Jiang Liu, Hui Ding, Zhaowei Cai, Yuting Zhang, Ravi~Kumar Satzoda, Vijay Mahadevan, and R Manmatha.
\newblock Polyformer: Referring image segmentation as sequential polygon generation.
\newblock In \emph{Proceedings of the IEEE/CVF Conference on Computer Vision and Pattern Recognition}, pages 18653--18663, 2023{\natexlab{c}}.

\bibitem[Loshchilov and Hutter(2017)]{loshchilov2017adamw}
Ilya Loshchilov and Frank Hutter.
\newblock Decoupled weight decay regularization.
\newblock \emph{arXiv preprint arXiv:1711.05101}, 2017.

\bibitem[L{\"u}ddecke and Ecker(2022)]{luddecke2022imagetextprompt}
Timo L{\"u}ddecke and Alexander Ecker.
\newblock Image segmentation using text and image prompts.
\newblock In \emph{Proceedings of the IEEE/CVF Conference on Computer Vision and Pattern Recognition}, pages 7086--7096, 2022.

\bibitem[L\"uddecke and Ecker(2022)]{lueddecke22clipseg}
Timo L\"uddecke and Alexander Ecker.
\newblock Image segmentation using text and image prompts.
\newblock In \emph{Proceedings of the IEEE/CVF Conference on Computer Vision and Pattern Recognition (CVPR)}, pages 7086--7096, 2022.

\bibitem[Milletari et~al.(2016)Milletari, Navab, and Ahmadi]{milletari2016dice}
Fausto Milletari, Nassir Navab, and Seyed-Ahmad Ahmadi.
\newblock V-net: Fully convolutional neural networks for volumetric medical image segmentation.
\newblock In \emph{2016 fourth international conference on 3D vision (3DV)}, pages 565--571. IEEE, 2016.

\bibitem[Min et~al.(2021)Min, Kang, and Cho]{min2021hsnet}
Juhong Min, Dahyun Kang, and Minsu Cho.
\newblock Hypercorrelation squeeze for few-shot segmentation.
\newblock In \emph{Proceedings of the IEEE/CVF International Conference on Computer Vision (ICCV)}, 2021.

\bibitem[Peng et~al.(2023)Peng, Wang, Dong, Hao, Huang, Ma, and Wei]{peng2023kosmos}
Zhiliang Peng, Wenhui Wang, Li Dong, Yaru Hao, Shaohan Huang, Shuming Ma, and Furu Wei.
\newblock Kosmos-2: Grounding multimodal large language models to the world.
\newblock \emph{arXiv preprint arXiv:2306.14824}, 2023.

\bibitem[Plummer et~al.(2017)Plummer, Wang, Cervantes, Caicedo, Hockenmaier, and Lazebnik]{flickrentities}
Bryan~A. Plummer, Liwei Wang, Christopher~M. Cervantes, Juan~C. Caicedo, Julia Hockenmaier, and Svetlana Lazebnik.
\newblock Flickr30k entities: Collecting region-to-phrase correspondences for richer image-to-sentence models.
\newblock \emph{IJCV}, 123\penalty0 (1):\penalty0 74--93, 2017.

\bibitem[Radford et~al.(2021)Radford, Kim, Hallacy, Ramesh, Goh, Agarwal, Sastry, Askell, Mishkin, Clark, et~al.]{radford2021clip}
Alec Radford, Jong~Wook Kim, Chris Hallacy, Aditya Ramesh, Gabriel Goh, Sandhini Agarwal, Girish Sastry, Amanda Askell, Pamela Mishkin, Jack Clark, et~al.
\newblock Learning transferable visual models from natural language supervision.
\newblock In \emph{International conference on machine learning}, pages 8748--8763. PMLR, 2021.

\bibitem[Radford et~al.(2023)Radford, Kim, Xu, Brockman, McLeavey, and Sutskever]{radford2023whisper}
Alec Radford, Jong~Wook Kim, Tao Xu, Greg Brockman, Christine McLeavey, and Ilya Sutskever.
\newblock Robust speech recognition via large-scale weak supervision.
\newblock In \emph{International Conference on Machine Learning}, pages 28492--28518. PMLR, 2023.

\bibitem[Rajbhandari et~al.(2020)Rajbhandari, Rasley, Ruwase, and He]{rajbhandari2020zero}
Samyam Rajbhandari, Jeff Rasley, Olatunji Ruwase, and Yuxiong He.
\newblock Zero: Memory optimizations toward training trillion parameter models.
\newblock In \emph{SC20: International Conference for High Performance Computing, Networking, Storage and Analysis}, pages 1--16. IEEE, 2020.

\bibitem[Ramanathan et~al.(2023)Ramanathan, Kalia, Petrovic, Wen, Zheng, Guo, Wang, Marquez, Kovvuri, Kadian, Mousavi, Song, Dubey, and Mahajan]{ramanathan2023paco}
Vignesh Ramanathan, Anmol Kalia, Vladan Petrovic, Yi Wen, Baixue Zheng, Baishan Guo, Rui Wang, Aaron Marquez, Rama Kovvuri, Abhishek Kadian, Amir Mousavi, Yiwen Song, Abhimanyu Dubey, and Dhruv Mahajan.
\newblock {PACO}: Parts and attributes of common objects.
\newblock In \emph{arXiv preprint arXiv:2301.01795}, 2023.

\bibitem[Rasley et~al.(2020)Rasley, Rajbhandari, Ruwase, and He]{rasley2020deepspeed}
Jeff Rasley, Samyam Rajbhandari, Olatunji Ruwase, and Yuxiong He.
\newblock Deepspeed: System optimizations enable training deep learning models with over 100 billion parameters.
\newblock In \emph{Proceedings of the 26th ACM SIGKDD International Conference on Knowledge Discovery \& Data Mining}, pages 3505--3506, 2020.

\bibitem[Tanaka et~al.(2019)Tanaka, Itamochi, Narioka, Sato, Ushiku, and Harada]{tanaka2019easytounderstand}
Mikihiro Tanaka, Takayuki Itamochi, Kenichi Narioka, Ikuro Sato, Yoshitaka Ushiku, and Tatsuya Harada.
\newblock Generating easy-to-understand referring expressions for target identifications.
\newblock In \emph{Proceedings of the IEEE/CVF International Conference on Computer Vision}, pages 5794--5803, 2019.

\bibitem[Touvron et~al.(2023)Touvron, Lavril, Izacard, Martinet, Lachaux, Lacroix, Rozi{\`e}re, Goyal, Hambro, Azhar, et~al.]{touvron2023llama}
Hugo Touvron, Thibaut Lavril, Gautier Izacard, Xavier Martinet, Marie-Anne Lachaux, Timoth{\'e}e Lacroix, Baptiste Rozi{\`e}re, Naman Goyal, Eric Hambro, Faisal Azhar, et~al.
\newblock Llama: Open and efficient foundation language models.
\newblock \emph{arXiv preprint arXiv:2302.13971}, 2023.

\bibitem[Vedantam et~al.(2015)Vedantam, Lawrence~Zitnick, and Parikh]{vedantam2015cider}
Ramakrishna Vedantam, C Lawrence~Zitnick, and Devi Parikh.
\newblock Cider: Consensus-based image description evaluation.
\newblock In \emph{Proceedings of the IEEE conference on computer vision and pattern recognition}, pages 4566--4575, 2015.

\bibitem[Wang et~al.(2023{\natexlab{a}})Wang, Chen, Chen, Wu, Zhu, Zeng, Luo, Lu, Zhou, Qiao, et~al.]{wang2023visionllm}
Wenhai Wang, Zhe Chen, Xiaokang Chen, Jiannan Wu, Xizhou Zhu, Gang Zeng, Ping Luo, Tong Lu, Jie Zhou, Yu Qiao, et~al.
\newblock Visionllm: Large language model is also an open-ended decoder for vision-centric tasks.
\newblock \emph{arXiv preprint arXiv:2305.11175}, 2023{\natexlab{a}}.

\bibitem[Wang et~al.(2023{\natexlab{b}})Wang, Shi, Li, Wang, Huang, Xing, Chen, Li, Zhu, Cao, et~al.]{wang2023asm}
Weiyun Wang, Min Shi, Qingyun Li, Wenhai Wang, Zhenhang Huang, Linjie Xing, Zhe Chen, Hao Li, Xizhou Zhu, Zhiguo Cao, et~al.
\newblock The all-seeing project: Towards panoptic visual recognition and understanding of the open world.
\newblock \emph{arXiv preprint arXiv:2308.01907}, 2023{\natexlab{b}}.

\bibitem[Wang et~al.(2023{\natexlab{c}})Wang, Wang, Cao, Shen, and Huang]{wang2023painter}
Xinlong Wang, Wen Wang, Yue Cao, Chunhua Shen, and Tiejun Huang.
\newblock Images speak in images: A generalist painter for in-context visual learning.
\newblock In \emph{Proceedings of the IEEE/CVF Conference on Computer Vision and Pattern Recognition}, pages 6830--6839, 2023{\natexlab{c}}.

\bibitem[Wang et~al.(2023{\natexlab{d}})Wang, Zhang, Cao, Wang, Shen, and Huang]{wang2023seggpt}
Xinlong Wang, Xiaosong Zhang, Yue Cao, Wen Wang, Chunhua Shen, and Tiejun Huang.
\newblock Seggpt: Segmenting everything in context.
\newblock \emph{arXiv preprint arXiv:2304.03284}, 2023{\natexlab{d}}.

\bibitem[Wang et~al.(2022)Wang, Lu, Li, Tao, Guo, Gong, and Liu]{wang2022cris}
Zhaoqing Wang, Yu Lu, Qiang Li, Xunqiang Tao, Yandong Guo, Mingming Gong, and Tongliang Liu.
\newblock Cris: Clip-driven referring image segmentation.
\newblock In \emph{Proceedings of the IEEE/CVF conference on computer vision and pattern recognition}, pages 11686--11695, 2022.

\bibitem[Wu et~al.(2020)Wu, Lin, Cohen, Bui, and Maji]{wu2020phrasecut}
Chenyun Wu, Zhe Lin, Scott Cohen, Trung Bui, and Subhransu Maji.
\newblock Phrasecut: Language-based image segmentation in the wild.
\newblock In \emph{Proceedings of the IEEE/CVF Conference on Computer Vision and Pattern Recognition}, pages 10216--10225, 2020.

\bibitem[Wu et~al.(2022)Wu, Wang, Yang, Gan, Liu, Yuan, and Wang]{wu2022grit}
Jialian Wu, Jianfeng Wang, Zhengyuan Yang, Zhe Gan, Zicheng Liu, Junsong Yuan, and Lijuan Wang.
\newblock Grit: A generative region-to-text transformer for object understanding.
\newblock \emph{arXiv preprint arXiv:2212.00280}, 2022.

\bibitem[Yan et~al.(2023)Yan, Jiang, Wu, Wang, Yuan, Luo, and Lu]{UNINEXT}
Bin Yan, Yi Jiang, Jiannan Wu, Dong Wang, Zehuan Yuan, Ping Luo, and Huchuan Lu.
\newblock Universal instance perception as object discovery and retrieval.
\newblock In \emph{CVPR}, 2023.

\bibitem[Yang et~al.(2022)Yang, Wang, Tang, Chen, Zhao, and Torr]{yang2022lavt}
Zhao Yang, Jiaqi Wang, Yansong Tang, Kai Chen, Hengshuang Zhao, and Philip~HS Torr.
\newblock Lavt: Language-aware vision transformer for referring image segmentation.
\newblock In \emph{CVPR}, 2022.

\bibitem[Ye et~al.(2023)Ye, Long, Feng, and Wang]{ye2023ireg}
Fulong Ye, Yuxing Long, Fangxiang Feng, and Xiaojie Wang.
\newblock Whether you can locate or not? interactive referring expression generation.
\newblock In \emph{Proceedings of the 31st ACM International Conference on Multimedia}, pages 4697--4706, 2023.

\bibitem[Yu et~al.(2016{\natexlab{a}})Yu, Poirson, Yang, Berg, and Berg]{yu2016modeling}
Licheng Yu, Patrick Poirson, Shan Yang, Alexander~C Berg, and Tamara~L Berg.
\newblock Modeling context in referring expressions.
\newblock In \emph{Computer Vision--ECCV 2016: 14th European Conference, Amsterdam, The Netherlands, October 11-14, 2016, Proceedings, Part II 14}, pages 69--85. Springer, 2016{\natexlab{a}}.

\bibitem[Yu et~al.(2016{\natexlab{b}})Yu, Poirson, Yang, Berg, and Berg]{yu2016refcocog}
Licheng Yu, Patrick Poirson, Shan Yang, Alexander~C Berg, and Tamara~L Berg.
\newblock Modeling context in referring expressions.
\newblock In \emph{Computer Vision--ECCV 2016: 14th European Conference, Amsterdam, The Netherlands, October 11-14, 2016, Proceedings, Part II 14}, pages 69--85. Springer, 2016{\natexlab{b}}.

\bibitem[Yu et~al.(2016{\natexlab{c}})Yu, Poirson, Yang, Berg, and Berg]{yu2016visdif}
Licheng Yu, Patrick Poirson, Shan Yang, Alexander~C Berg, and Tamara~L Berg.
\newblock Modeling context in referring expressions.
\newblock In \emph{Computer Vision--ECCV 2016: 14th European Conference, Amsterdam, The Netherlands, October 11-14, 2016, Proceedings, Part II 14}, pages 69--85. Springer, 2016{\natexlab{c}}.

\bibitem[Yu et~al.(2017)Yu, Tan, Bansal, and Berg]{yu2017slr}
Licheng Yu, Hao Tan, Mohit Bansal, and Tamara~L Berg.
\newblock A joint speaker-listener-reinforcer model for referring expressions.
\newblock In \emph{Proceedings of the IEEE conference on computer vision and pattern recognition}, pages 7282--7290, 2017.

\bibitem[Zhang et~al.(2022)Zhang, Sun, Yang, and Chen]{zhang2022fptrans}
Jian-Wei Zhang, Yifan Sun, Yi Yang, and Wei Chen.
\newblock Feature-proxy transformer for few-shot segmentation.
\newblock \emph{Advances in Neural Information Processing Systems}, 35:\penalty0 6575--6588, 2022.

\bibitem[Zhang et~al.(2023{\natexlab{a}})Zhang, Han, Liu, Gao, Zhou, Hu, Yan, Lu, Li, and Qiao]{zhang2023llamaadapter}
Renrui Zhang, Jiaming Han, Chris Liu, Peng Gao, Aojun Zhou, Xiangfei Hu, Shilin Yan, Pan Lu, Hongsheng Li, and Yu Qiao.
\newblock Llama-adapter: Efficient fine-tuning of language models with zero-init attention.
\newblock \emph{arXiv preprint arXiv:2303.16199}, 2023{\natexlab{a}}.

\bibitem[Zhang et~al.(2023{\natexlab{b}})Zhang, Sun, Chen, Xiao, Shao, Zhang, Chen, and Luo]{zhang2023gpt4roi}
Shilong Zhang, Peize Sun, Shoufa Chen, Min Xiao, Wenqi Shao, Wenwei Zhang, Kai Chen, and Ping Luo.
\newblock Gpt4roi: Instruction tuning large language model on region-of-interest, 2023{\natexlab{b}}.

\bibitem[Zhao et~al.(2023)Zhao, Lin, Zhou, Huang, Feng, and Kang]{zhao2023bubogpt}
Yang Zhao, Zhijie Lin, Daquan Zhou, Zilong Huang, Jiashi Feng, and Bingyi Kang.
\newblock Bubogpt: Enabling visual grounding in multi-modal llms.
\newblock \emph{arXiv preprint arXiv:2307.08581}, 2023.

\bibitem[Zhou et~al.(2019)Zhou, Zhao, Puig, Xiao, Fidler, Barriuso, and Torralba]{zhou2019ade20k}
Bolei Zhou, Hang Zhao, Xavier Puig, Tete Xiao, Sanja Fidler, Adela Barriuso, and Antonio Torralba.
\newblock Semantic understanding of scenes through the ade20k dataset.
\newblock \emph{International Journal of Computer Vision}, 127:\penalty0 302--321, 2019.

\bibitem[Zhou et~al.(2022)Zhou, Wang, Zhang, Sun, Zhang, Birchfield, Guo, Kong, Wang, and Zhong]{zhou2022avs}
Jinxing Zhou, Jianyuan Wang, Jiayi Zhang, Weixuan Sun, Jing Zhang, Stan Birchfield, Dan Guo, Lingpeng Kong, Meng Wang, and Yiran Zhong.
\newblock Audio--visual segmentation.
\newblock In \emph{European Conference on Computer Vision}, pages 386--403. Springer, 2022.

\bibitem[Zhu et~al.(2023)Zhu, Chen, Shen, Li, and Elhoseiny]{zhu2023minigpt4}
Deyao Zhu, Jun Chen, Xiaoqian Shen, Xiang Li, and Mohamed Elhoseiny.
\newblock Minigpt-4: Enhancing vision-language understanding with advanced large language models.
\newblock \emph{arXiv preprint arXiv:2304.10592}, 2023.

\bibitem[Zou et~al.(2023{\natexlab{a}})Zou, Dou, Yang, Gan, Li, Li, Dai, Behl, Wang, Yuan, et~al.]{zou2023xdecoder}
Xueyan Zou, Zi-Yi Dou, Jianwei Yang, Zhe Gan, Linjie Li, Chunyuan Li, Xiyang Dai, Harkirat Behl, Jianfeng Wang, Lu Yuan, et~al.
\newblock Generalized decoding for pixel, image, and language.
\newblock In \emph{Proceedings of the IEEE/CVF Conference on Computer Vision and Pattern Recognition}, pages 15116--15127, 2023{\natexlab{a}}.

\bibitem[Zou et~al.(2023{\natexlab{b}})Zou, Yang, Zhang, Li, Li, Gao, and Lee]{zou2023seem}
Xueyan Zou, Jianwei Yang, Hao Zhang, Feng Li, Linjie Li, Jianfeng Gao, and Yong~Jae Lee.
\newblock Segment everything everywhere all at once.
\newblock \emph{arXiv preprint arXiv:2304.06718}, 2023{\natexlab{b}}.

\end{thebibliography}
}


\end{document}